\documentclass{article}
\usepackage[preprint]{neurips_2026}
\usepackage[utf8]{inputenc}
\usepackage[T1]{fontenc}
\usepackage{hyperref}
\usepackage{url}
\usepackage{booktabs}
\usepackage{arydshln}
\usepackage{amsfonts}
\usepackage{nicefrac}
\usepackage{microtype}
\usepackage{xcolor}
\usepackage{placeins}
\usepackage{longtable}
\usepackage{graphicx}
\usepackage{pgffor}
\usepackage{amsmath,amssymb,amsthm}
\usepackage{graphicx}
\usepackage{subcaption}
\usepackage{algorithm}
\usepackage{algorithmic}
\usepackage{cite}
\usepackage{array}
\usepackage{multirow}
\usepackage{bm}
\usepackage{mathtools}
\usepackage{natbib}
\usepackage{wrapfig}
\usepackage{nicefrac}
\usepackage{bm}
\usepackage{tikz}
\newcommand{\std}[1]{\textcolor{gray}{\scriptsize\,$\pm$ #1}}
\newcommand{\zoomimage}[5]{
\begin{tikzpicture}
    \node[anchor=south west, inner sep=0] (img) at (0,0)
        {\includegraphics[width=#2]{#1}};
    \node[
        anchor=south east,
        draw=red,
        line width=0.3pt,
        inner sep=0pt,
        #5
    ] at (img.south east) {
        \includegraphics[
            width=#4,
            trim=#3,
            clip
        ]{#1}
    };
\end{tikzpicture}
}
\newcommand{\x}{\boldsymbol{x}}
\newcommand{\y}{\boldsymbol{y}}
\newcommand{\K}{\boldsymbol{K}}
\newcommand{\D}{\boldsymbol{D}}
\newcommand{\I}{\boldsymbol{I}}
\newcommand{\eps}{\bm{\epsilon}}

\newcommand{\E}{\mathbb{E}}

\usepackage[dvipsnames]{xcolor}

\newtheorem{proposition}{Proposition}

\title{Improving Diffusion Posterior Samplers with Lagged Temporal Corrections for Image Restoration}
\author{
  Davide Evangelista \\
  Dept. of Computer Science and Engineering\\
  University of Bologna \\
  \texttt{davide.evangelista5@unibo.it}
  \And
  Elena Morotti \\
  Dept. of Political and Social Sciences\\
  University of Bologna \\
  \texttt{elena.morotti4@unibo.it}
  \AND
  Francesco Pivi \\
  Dept. of Computer Science and Engineering \\
  University of Bologna \\
  \texttt{francesco.pivi2@unibo.it}
  \And
  Maurizio Gabbrielli \\
  Dept. of Computer Science and Engineering\\
  University of Bologna \\
  \texttt{maurizio.gabbrielli@unibo.it}
}
\begin{document}
\maketitle
\begin{abstract}
Diffusion-based posterior sampling (PS) is a leading framework for imaging inverse problems, combining learned priors with measurement constraints. Yet, its standard formulations rely on instantaneous data-consistent estimates, which induce temporal variability in the reverse dynamics. We reinterpret PS from a dynamical perspective, showing that the standard PS update corresponds to a first-order discretization of the diffusion dynamics plus a residual correction capturing the mismatch between the denoised prediction and the data-consistent estimate. A second-order discretization, however, naturally introduces a temporal correction based on the variation of consecutive estimates. \\
Building on this, we propose LAMP, combining the second-order update with the residual correction characterizing a PS technique. LAMP thus inherits a lagged temporal correction, and it can be implemented as a modular plug-in over the PS backbone. We show that LAMP preserves the structure of a posterior sampler, and we perform a one-step risk analysis to characterize when LAMP improves the reverse transition via a bias-variance trade-off.
Experiments across multiple imaging tasks demonstrate consistent improvements over strong baselines such as DiffPIR and DDRM, without increasing the number of denoising evaluations.
\end{abstract}
\section{Introduction}\label{sec:introduction}
Inverse problems are prevalent in imaging, with applications in restoration, computational photography, and medical reconstruction. The objective is to recover an unknown signal $\x_0$ from measurements $\y = \K \x_0 + \boldsymbol{e}$, where $\K$ is a possibly ill-conditioned forward operator and $\boldsymbol{e} \sim \mathcal{N}(0,\sigma_{\y}^2 \I)$ represents noise corruption.
Classical approaches rely on variational methods, which cast the task as an optimization problem regularized by hand-crafted priors. While effective in many settings, these models often struggle to capture complex image statistics, especially in highly ill-posed regimes.
Deep learning has emerged as a powerful alternative, enabling significant improvements in reconstruction quality by learning data-driven priors from large datasets. In particular, diffusion-based generative models have demonstrated remarkable performance in modeling complex high-dimensional data distributions \citep{ho2020ddpm,song2021scorebased}. Building on this success, recent works have proposed diffusion models as powerful priors for solving inverse problems, achieving state-of-the-art results across a range of imaging tasks. These approaches reinterpret image processing as a \emph{posterior sampling} (PS) task: given a forward model and noisy measurements, the goal is to sample from the posterior distribution of the unknown image conditioned on the observations. In this framework, a pre-trained diffusion model acts as a powerful prior, while data consistency is enforced during the sampling process.
More precisely, let $\x_t^{\mathrm{PS}}$ denote the current iterate at diffusion time $t$. Each standard posterior sampling update combines a denoising step provided by the diffusion model with a correction enforcing consistency with the observed data. This leads to updates of the general form:
\begin{equation}\label{eq:generic_posterior_bg}
\x_{t-\Delta t}^{\mathrm{PS}} = \alpha_{t-\Delta t} \D_{t-\Delta t} + \sigma_{t-\Delta t} \eps_\Theta(\x_t^{\mathrm{PS}},t),
\end{equation}
where $\{ \alpha_t \}_{t=0}^{T-1}$ and $\{ \sigma_t \}_{t=0}^{T-1}$ denote the time-dependent signal and noise coefficients satisfying $\alpha_t^2 + \sigma_t^2 = 1$ for any $t$, $\Delta t > 0$ is the time discretization term, $\eps_\Theta(\cdot,\cdot)$ is the learned noise predictor, and $\D_t$ is a data-consistent estimate.
Variants of this approach underlie several recent methods, which differ in how the data-consistency step is implemented and integrated within the diffusion dynamics. For instance, $\D_t$ is obtained via a proximal step in the DiffPIR method~\citep{zhu2023diffpir}, via a projection onto the measurement constraints in the DDNM method~\citep{wang2023ddnm}, or in closed form in a transformed domain when the forward operator admits a suitable decomposition in the DDRM method~\citep{kawar2022ddrm}. More details are provided in Appendix \ref{app:dt_diffpir_ddrm}.
In all cases, existing PS methods rely on instantaneous, step-wise data-consistent estimates $\D_t$ that are recomputed and injected at each reverse iteration.
This design introduces instability in the correction term, as small changes in the iterate $\x_t$ can lead to inconsistent updates due to denoising error, discretization, and model mismatch. As a result, the reverse dynamics are effectively driven by a time-varying and potentially noisy forcing signal.
\begin{figure}[t]
    \centering
    \begin{minipage}[t]{0.40\textwidth}
        \vspace{0pt}
        \centering
        \fbox{\includegraphics[width=0.9\linewidth]{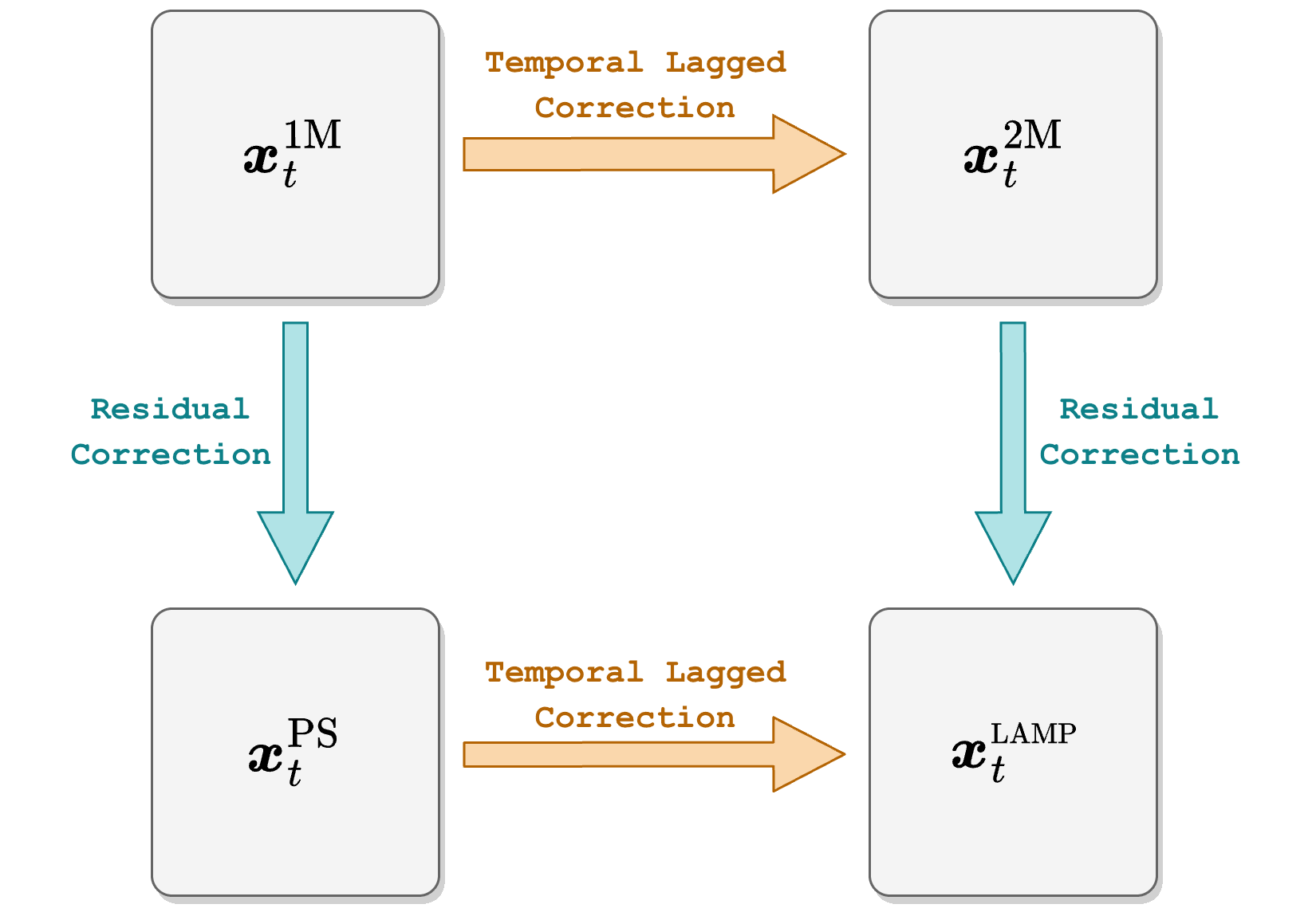}}
    \end{minipage}
    \hfill
    \begin{minipage}[t]{0.55\textwidth}
        \vspace{0pt}
    \scriptsize
    \setlength{\tabcolsep}{1pt} \vspace{-2mm}
    \begin{tabular}{@{}cc@{\hspace{2pt}}c@{\hspace{2pt}}c@{\hspace{2pt}}c@{}}
        &Measurement & DDRM & DDRM-LAMP & Ground-truth \\
        \rotatebox{90}{\hspace{1px}Motion Blur}&
        \zoomimage
        {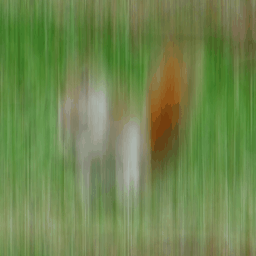}
        {0.22\linewidth}
        {80 120 120 80}
        {0.08\linewidth}
        {}
        &
        \zoomimage
        {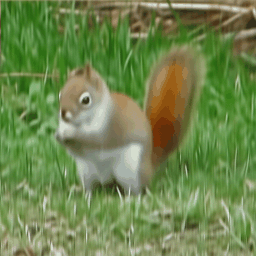}
        {0.22\linewidth}
        {80 120 120 80}
        {0.08\linewidth}
        {}
        &
        \zoomimage
        {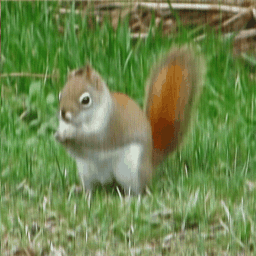}
        {0.22\linewidth}
        {80 120 120 80}
        {0.08\linewidth}
        {}
        &
        \zoomimage
        {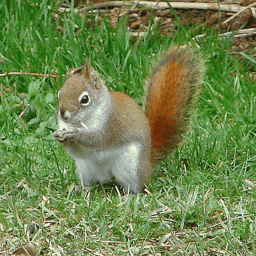}
        {0.22\linewidth}
        {80 120 120 80}
        {0.08\linewidth}
        {}
        \\
        \rotatebox{90}{\hspace{1px}Gaussian Blur}&
        \zoomimage
        {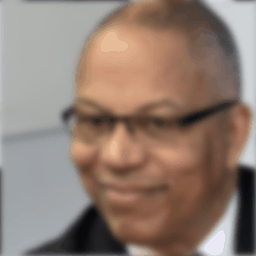}
        {0.22\linewidth}
        {100 20 100 180}
        {0.08\linewidth}
        {}
        &
        \zoomimage
        {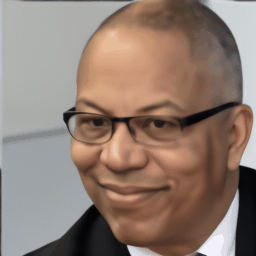}
        {0.22\linewidth}
        {100 20 100 180}
        {0.08\linewidth}
        {}
        &
        \zoomimage
        {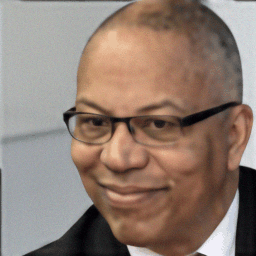}
        {0.22\linewidth}
        {100 20 100 180}
        {0.08\linewidth}
        {}
        &
        \zoomimage
        {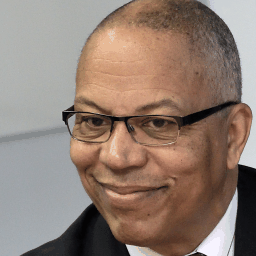}
        {0.22\linewidth}
        {100 20 100 180}
        {0.08\linewidth}
        {}
        \\
    \end{tabular}
    \end{minipage}
    \caption{Overview of the LAMP scheme.
    Left: relation among the updates of the considered diffusive samplers for a fixed time. Right: examples showing LAMP onto the DDRM method.}
    \label{fig:Quadrato_FirstResults}
\end{figure}
From a dynamical perspective, this leads to two key limitations. First, the resulting trajectory is not temporally smooth, deviating from the behavior expected from a well-defined continuous-time process. Second, the method lacks temporal memory, as each correction $\D_t$ is computed independently despite the strong correlation between consecutive estimates. These effects are largely overlooked in current approaches, yet they play a crucial role in the stability and consistency of diffusion-based inverse solvers.
This lack of temporal coherence also reveals a fundamental mismatch with recent advances in diffusion PS for unconditional generation based on probability-flow ODEs \citep{song2021scorebased}. In this context, in fact, high-order solvers such as DPM-Solver \citep{lu2022dpmsolver++} achieve remarkable acceleration by assuming that the reverse process follows a smooth and well-defined continuous-time dynamics. However, in the presence of measurement constraints, this assumption no longer holds. The data-consistency corrections introduced by PS act as a time-dependent forcing term that perturbs the reverse dynamics at every step. As a result, the trajectory deviates from that of a standard probability-flow ODE, both in terms of smoothness and consistency, limiting the direct applicability of classical high-order solvers to inverse problems.
Motivated by this perspective, our contributions are as follows:
\begin{itemize}
    \item We identify a structural mismatch between ODE-based diffusion solvers and diffusive PS methods for inverse problems, showing that the latter are implicitly driven by a residual-dependent forcing term that is not captured by standard probability-flow ODE formulations.
    \item We reinterpret posterior samplers as discrete dynamical systems and show that standard posterior updates correspond to a first-order discretization of a diffusion dynamics corrected by a residual-dependent force. Building on this perspective, we propose \textbf{LAMP} (LAgged Multistep Posterior) as a second-order corrected scheme, explicitly accounting for both the temporal evolution and a residual-dependent term, as sketched in the first panel of Figure \ref{fig:Quadrato_FirstResults}.
    \item From a theoretical perspective, we show that LAMP preserves the standard posterior sampling structure in Eq.~\eqref{eq:generic_posterior_bg}, as its update can be expressed in the same form with a modified data-consistent estimate that incorporates temporal information. This establishes LAMP as a modular extension of PS methods, rather than a separate reconstruction scheme.
    \item We show that LAMP can be instantiated on top of \emph{any} posterior sampling (PS) method, acting as a plug-in mechanism that can be layered on top of an arbitrary choice of $\D_t$. Moreover, we prove that the resulting method remains itself a valid PS method, thereby ensuring full compatibility with the underlying posterior sampling framework.
    \item We provide a one-step risk analysis showing that, under a mild condition, the error induced by LAMP is smaller than that of the base PS method on which it is built. We further show that this condition is typically satisfied in the later stages of the diffusion process, which motivates a hybrid strategy where early iterations follow the base PS method, while LAMP is activated in later steps to improve reconstruction quality and stability.
\end{itemize}
\section{Related Works}
\label{sec:related_work}
\textbf{Diffusion models for image generation.}
Diffusion probabilistic models have become a standard paradigm for generative modeling, starting from DDPM \citep{ho2020ddpm} and score-based formulations \citep{song2021scorebased,sohl2015deep}. These models admit both stochastic (SDE) and deterministic (ODE) interpretations, the latter leading to probability-flow dynamics that can be discretized for sampling \citep{song2021scorebased,song2021ddim}. This connection has motivated a large body of work on accelerating sampling through improved numerical solvers, including DDIM \citep{song2021ddim}, PNDM \citep{liu2022pndm}, DPM-Solver \citep{lu2022dpmsolver++}, and subsequent higher-order and accelerated methods \citep{lu2022dpmsolver++,zhang2023fast}. These approaches are derived under the assumption that the reverse process follows an underlying ODE, and therefore aim to reduce the discretization error of that continuous dynamics. In contrast, our work highlights that such ODE-based interpretations do not fully capture the behavior of posterior sampling methods for inverse problems, where additional residual-dependent terms arise.
\textbf{Diffusion models for inverse problems.}
Recently, diffusion models have been adapted to inverse problems by integrating measurement constraints directly into the reverse dynamics. Early approaches combine pretrained diffusion models with projection or conditioning strategies \citep{kawar2021snips,whang2022deblurring}. A large class of methods can be interpreted as posterior sampling (PS) procedures, where data consistency is enforced at each step through an explicit update. MCG introduces a manifold-constrained gradient to enforce measurement consistency while maintaining proximity to the data manifold \citep{chung2022improving}. DDRM formulates posterior sampling for linear inverse problems via a variational approximation, achieving strong performance and efficiency \citep{kawar2022ddrm}. DPS generalizes posterior sampling to noisy and nonlinear forward models through a Laplace approximation \citep{chung2022dps}. DDNM instead exploits a range/null-space decomposition to enforce measurement consistency in linear settings, enabling zero-shot restoration without retraining \citep{wang2023ddnm}. Related approaches also include plug-and-play diffusion priors, where a diffusion denoiser is combined with iterative data-consistency updates, as in DiffPIR \citep{zhu2023diffpir}. While these methods differ in how the posterior update is constructed, they all share a common iterative structure, which we reinterpret in this work as a discretization of an underlying diffusion dynamics with an additional residual-dependent forcing term.
\textbf{Positioning of our method.}
Rather than introducing a new posterior update, prior, or conditioning strategy, our contribution focuses on the temporal structure of existing PS methods. We identify a previously underexplored source of instability, namely the temporal variability of measurement-aware estimates across iterations, and show that standard PS updates correspond to a first-order discretization of a forced diffusion dynamics. This perspective naturally leads to a multistep correction scheme that can be applied on top of any PS method, yielding a modular improvement without modifying the underlying model or training procedure.
\section{Methodology}
\label{sec:methodology}
In this section, we derive LAMP as a multistep scheme based on a second-order discretization of the diffusion dynamics with a residual-correction term. More details, including proofs of the theoretical results, are provided in Appendix \ref{app:dt_diffpir_ddrm} and \ref{app:proofs}.
We first recall the minimal DDIM and probability-flow identities needed to fix notation.
\subsection{Background}
Consider a fixed timestep $t$. Using the variance-preserving parameterization $\alpha_t^2+\sigma_t^2=1$, and the standard marginal noising relation $\x_t = \alpha_t \x_0 + \sigma_t \eps_t$, with $\eps_t\sim\mathcal{N}(0,\I)$, the clean-image Tweedie estimation \citep{tweedie1957statistical,ho2020ddpm} given the noise predictor $\eps_\Theta(\x_t,t)$ reads
\begin{equation}
    \hat{\x}_{0|t} := \mathbb{E}\left[ \x_0 | \x_t\right]
    =
    \frac{\x_t-\sigma_t\eps_\Theta(\x_t,t)}{\alpha_t}.
    \label{eq:x0hat_method}
\end{equation}
DDIM \citep{song2021ddim} then defines the deterministic updates at time $t$ as
\begin{equation}
    \x_{t-\Delta t}
    =
    \alpha_{t-\Delta t}\hat{\x}_{0|t}
    +
    \sigma_{t-\Delta t}\eps_\Theta(\x_t,t).
    \label{eq:ddim_discrete_method}
\end{equation}
A complementary line of work interprets the generation process as the discretization of a continuous-time probability-flow ordinary differential equation (PF-ODE) \citep{song2021scorebased}. In the variance-preserving setting, this ODE can be written directly in terms of the signal and noise rates $\alpha_t$ and $\sigma_t$ as
\begin{equation}
    \frac{d\x_t}{dt}
    =
    \frac{d\log \sigma_t}{dt}\,\x_t
    +
    \frac{d}{dt}\left(\frac{\alpha_t}{\sigma_t}\right)
    \sigma_t^2 \hat{\x}_{0|t},
    \label{eq:PFODE_alpha_sigma}
\end{equation}
where $\hat{\x}_{0|t}$ is defined in \eqref{eq:x0hat_method}. It is often convenient to reparameterize time using the log-SNR variable $\lambda_t := \log(\alpha_t / \sigma_t)$, which provides a more uniform parametrization of the diffusion trajectory. Using this change of variables, Eq.~\eqref{eq:PFODE_alpha_sigma} can be rewritten in integrating-factor form as
\begin{equation}
    \frac{d}{d\lambda}
    \left(
        \frac{\x(\lambda)}{\sigma(\lambda)}
    \right)
    =
    \frac{\alpha(\lambda)}{\sigma(\lambda)}
    \hat{\x}_0(\x(\lambda),\lambda),
    \label{eq:pf_ode_if_method}
\end{equation}
which highlights a linear structure in the transformed variable $\x(\lambda)/\sigma(\lambda)$. Integrating from $\lambda_t$ to $\lambda_{t-\Delta t}$ and setting $h=\lambda_{t-\Delta t}-\lambda_t>0$, yields
\begin{equation}
    \x_{t-\Delta t}
    =
    \frac{\sigma_{t-\Delta t}}{\sigma_t}\x_t
    +
    \sigma_{t-\Delta t}
    \int_{\lambda_t}^{\lambda_{t-\Delta t}}
    \frac{\alpha(\lambda)}{\sigma(\lambda)}
    \hat{\x}_0(\x(\lambda),\lambda)
    \,d\lambda.
    \label{eq:mild_ddim_method}
\end{equation}
Approximating $\hat{\x}_0(\x(\lambda),\lambda)$ by the constant value $\hat{\x}_{0|t}$ over the interval $[\lambda_{t},\lambda_{t-\Delta t}]$, \eqref{eq:mild_ddim_method} becomes:
\begin{equation}
    \x_{t-\Delta t}
    =
    \frac{\sigma_{t-\Delta t}}{\sigma_t}\x_t
    +
    \alpha_{t-\Delta t} A_0(h)\hat{\x}_{0|t},
    \quad \text{with } A_0(h)=1-e^{-h} \approx h.
    \label{eq:ddim_1m_method}
\end{equation}
As proved in Appendix \ref{app:proof_ddim_1m}, this expression is algebraically equivalent to \eqref{eq:ddim_discrete_method}, implying that standard DDIM step can be interpreted as the first-order exponential discretization of the probability-flow ODE.  The idea of higher-order solvers such as DPM-Solver++ \citep{lu2022dpm}, is to achieve improved accuracy by approximating the integral in \eqref{eq:mild_ddim_method} using multistep polynomial expansions of the model prediction. In particular, employing 2-step approximation of the integrating factor $\hat{\x}_0(\x(\lambda),\lambda)$ leads to:
\begin{equation}\label{eq:2m-generation}
        \x_{t-\Delta t}
    =
    \frac{\sigma_{t-\Delta t}}{\sigma_t}\x_t
    +
    \alpha_{t-\Delta t} A_0(h)\hat{\x}_{0|t} + \alpha_{t - \Delta t} A_1(h)\gamma
    \frac{\hat{\x}_{0|t}-\hat{\x}_{0|t+\Delta t}}{\lambda_t-\lambda_{t+\Delta t}},
\end{equation}
where $A_1(h)$ is the standard second-order exponential coefficient, given by $A_1(h)=1-\frac{1-e^{-h}}{h} \approx \frac{h^2}{2}$.
\subsection{LAMP: Lagged Multistep Posterior Sampler}
When diffusive PS models are employed to solve inverse problems as $\y = \K\x_0+\boldsymbol{e}$, at each reverse step, the raw denoised estimate $\hat{\x}_{0|t}$ in \eqref{eq:ddim_discrete_method} is replaced with the measurement-aware estimate $\D_t$, and the reverse step takes the common form reported in \eqref{eq:generic_posterior_bg}. Here $\D_t = \mathcal{C}_t(\hat{\x}_{0|t},\y,\K)$ where $\mathcal{C}_t$ represents the image operator induced by the chosen posterior sampler, as already discussed in Section \ref{sec:introduction}.
A continuous analogue of \eqref{eq:generic_posterior_bg} can be obtained by replacing $\hat{\x}_0(\x(\lambda),\lambda)$ with a measurement-aware target $\D_\lambda := \D_{\lambda^{-1} \left( \lambda_t \right)}$ in \eqref{eq:pf_ode_if_method}:
\begin{equation}
    \frac{d}{d\lambda}
    \left(
        \frac{\x(\lambda)}{\sigma(\lambda)}
    \right)
    =
    \frac{\alpha(\lambda)}{\sigma(\lambda)}
    \D_\lambda,
    \label{eq:posterior_continuous_method}
\end{equation}
which, approximating $\D_\lambda$ by the constant value $\D_t$ on $[\lambda_t,\lambda_{t - \Delta t}]$, yield to the corresponding first-order exponential step:
\begin{equation}
    \x_{t-\Delta t}^{1\mathrm{M}}
    =
    \frac{\sigma_{t-\Delta t}}{\sigma_t}\x_t
    +
    \alpha_{t-\Delta t} A_0(h)\D_t .
    \label{eq:posterior_1m_method}
\end{equation}
However, unlike in the unconditional setting where DDIM coincides with a first-order exponential discretization, the posterior update rule in \eqref{eq:generic_posterior_bg} is not equal to \eqref{eq:posterior_1m_method}. Indeed, substituting $\eps_\Theta(\x_t,t) = \frac{\x_t-\alpha_t\hat{\x}_{0|t}}{\sigma_t}$ into \eqref{eq:generic_posterior_bg} gives the exact decomposition
\begin{equation}
    \x_{t-\Delta t}^{\mathrm{PS}}
    =
    \underbrace{\frac{\sigma_{t-\Delta t}}{\sigma_t}\x_t
    +
    \alpha_{t-\Delta t} A_0(h)\D_t}_{x_{t-\Delta t}^{1\mathrm{M}}}
    +
    \alpha_{t-\Delta t} e^{-h}
    \left(
        \D_t-\hat{\x}_{0|t}
    \right).
    \label{eq:posterior_PS_1M_forcing_method}
\end{equation}
Here, the first two terms correspond to the first-order exponential step associated with \eqref{eq:posterior_continuous_method}, i.e. to $\x_{t-\Delta t}^{1\mathrm{M}}$, while the last term is a \emph{residual} correction term, as it explicitly depends on the mismatch $\D_t - \hat{\x}_{0|t}$ between the data-consistent estimate and the diffusion prior. This term therefore captures the discrepancy between the two sources of information and acts as a corrective forcing that steers the reverse dynamics toward data consistency. In particular, it vanishes only when $\D_t = \hat{\x}_{0|t}$, i.e., when the prior prediction is already consistent with the measurements. As we will further illustrate in Section~\ref{sec:experiments}, this residual correction plays a beneficial role, leading to improved performance of $\x_{t-\Delta t}^{\mathrm{PS}}$ over its baseline first-order step $\x_{t-\Delta t}^{1\mathrm{M}}$.
In parallel, a second-order approximation of \eqref{eq:posterior_continuous_method} involves the temporal variation of the data-consistent estimates across two consecutive reverse steps. In particular, introducing $h_{\mathrm{prev}} := \lambda_t-\lambda_{t+\Delta t}$, a standard 2-step DPM-Solver-like discretization includes a \emph{temporal} term proportional to the discrete derivative $(\D_t-\D_{t+\Delta t})/h_{\mathrm{prev}}$, yielding
\begin{equation}
    \x_{t - \Delta t}^{2\mathrm{M}}
    =
    \x_{t - \Delta t}^{1\mathrm{M}}
    +
    \alpha_{t - \Delta t} A_1(h)\gamma
    \frac{\D_t-\D_{t + \Delta t}}{h_{\mathrm{prev}}},
    \label{eq:posterior_2M_1M}
\end{equation}
which corresponds to the posterior sampling in \eqref{eq:2m-generation}.
Motivated by the favorable role of the residual correction term linking $\x_{t - \Delta t}^{1\mathrm{M}}$ and $\x_{t-\Delta t}^{\mathrm{PS}}$, we introduce \textbf{LAMP} (LAgged Multistep Posterior), a multistep posterior sampler that combines the second-order discretization of the diffusion dynamics with the residual correction term. Building on \eqref{eq:posterior_PS_1M_forcing_method}, the LAMP update is defined as
\begin{equation}
    \x_{t - \Delta t}^{\mathrm{LAMP}}
    :=
    \x_{t - \Delta t}^{2\mathrm{M}}
    +
    \alpha_{t-\Delta t} e^{-h}
    \left(
        \D_t-\hat{\x}_{0|t}
    \right),
    \label{eq:lamp_def}
\end{equation}
i.e., as the two-step scheme derived from a second-order discretization of the diffusion dynamics boosted by the residual correction term. This idea is depicted in Figure \ref{fig:Quadrato_FirstResults}.
By combining \eqref{eq:posterior_PS_1M_forcing_method} and \eqref{eq:posterior_2M_1M}  into \eqref{eq:lamp_def}, we obtain the following update:
\begin{equation}
    \x_{t - \Delta t}^{\mathrm{LAMP}}
    =
    \x_{t - \Delta t}^{\mathrm{PS}}
    +
    \alpha_{t - \Delta t} A_1(h)\gamma
    \frac{\D_t-\D_{t + \Delta t}}{h_{\mathrm{prev}}},
    \label{eq:lamp_from_PS_method}
\end{equation}
which highlights the presence of the temporal correction term modulated by the coefficient $\gamma$. Precisely, the term $(\D_t-\D_{t + \Delta t})/h_{\mathrm{prev}}$ captures the variation of the data-consistent estimates across consecutive reverse steps, and can be interpreted as a discrete temporal derivative. The sign of $\gamma$ determines whether this correction extrapolates the sequence of estimates forward or introduces a stabilizing effect. In particular, choosing $\gamma < 0$ yields a \emph{lagged} temporal correction, where the current estimate is slightly pulled toward previous ones rather than extrapolated. This mechanism counteracts the variability induced by the instantaneous posterior correction, leading to smoother and more stable reverse dynamics. This motivates the name
LAMP, which emphasizes the role of lagged multistep information in the posterior update.
The following identity, whose proof is given in Appendix \ref{app:proof_lagged_form}, shows that the LAMP scheme preserves the posterior sampling structure.
\begin{proposition}[Lagged posterior form]
\label{prop:lamp_lagged_form}
Let $\beta_t:= -\gamma\frac{A_1(h)}{h_{\mathrm{prev}}} > 0$. Then, the LAMP update \eqref{eq:lamp_from_PS_method} can be written as
\begin{equation}
    \x_{t - \Delta t}^{\mathrm{LAMP}}
    =
    \alpha_{t - \Delta t}\widetilde{\D}_t
    +
    \sigma_{t - \Delta t}\eps_\Theta(\x_t,t),
    \label{eq:lamp_posterior_form_method}
\end{equation}
where
\begin{equation}
    \widetilde{\D}_t
    =
    (1-\beta_t)\D_t+\beta_t\D_{{t + \Delta t}}.
    \label{eq:lamp_filtered_target_method}
\end{equation}
\end{proposition}
Proposition~\ref{prop:lamp_lagged_form} is the key structural property of LAMP: the method is not an external post-processing step and it is not a classical higher-order solver for the unconditional probability-flow ODE. It is again a posterior sampler of the form \eqref{eq:generic_posterior_bg}, but with the instantaneous estimate $\D_t$ replaced by the temporally filtered estimate $\widetilde{\D}_t$. Therefore, LAMP can be applied on top of different correction operators $\mathcal{C}_t$ without changing the denoiser, the forward model, or the reverse schedule characterising state-of-the-art PS models.
We now give a local error argument explaining when the lagged estimate improves the reverse transition. First of all, note that although the LAMP trajectory is deterministic after initialization, the corrected estimate $\D_t$ is not exact, since it inherits denoising error, discretization error, and model mismatch. In the following, we rely on the assumption that $\D_t = \mu_t+\eta_t$, $\D_{{t + \Delta t}}=\mu_{t + \Delta t}+\eta_{t + \Delta t}$, where $\mu_t$ and $\mu_{{t + \Delta t}}$ are local posterior targets and $\eta_t,\eta_{t + \Delta t}$ are zero-mean estimation errors, such that $\E[\eta_t]=\E[\eta_{t + \Delta t}]=0$, $\mathrm{Var}(\eta_t)=\mathrm{Var}(\eta_{t + \Delta t})=\Sigma_t$, and $\mathrm{Cov}(\eta_t,\eta_{t + \Delta t})=\rho_t\Sigma_t$, with $0\leq \rho_t<1$. In Appendix \ref{app:proof_unequal_variances} we show a more general result that does not rely on this assumption.
\begin{proposition}[One-step risk comparison]
\label{prop:lamp_risk_method}
Let $\x_{t - \Delta t}^\mu := \alpha_{t - \Delta t}\mu_t+\sigma_{t - \Delta t}\eps_\Theta(\x_t,t)$ be the ideal one-step state obtained if the corrected estimate were equal to the local target $\mu_t$, and define $r_t:=\mu_{t + \Delta t}-\mu_t$. It holds:
\begin{equation}
    \E\|\x_{t - \Delta t}^{\mathrm{PS}}-\x_{t - \Delta t}^\mu\|^2
    =
    \alpha_{t - \Delta t}^2\mathrm{tr}(\Sigma_t),
    \label{eq:base_risk_method}
\end{equation}
whereas
\begin{equation}
    \E\|\x_{t - \Delta t}^{\mathrm{LAMP}}-\x_{t - \Delta t}^\mu\|^2
    =
    \alpha_{t - \Delta t}^2
    \left[
        \left(
            1-2\beta_t(1-\beta_t)(1-\rho_t)
        \right)
        \mathrm{tr}(\Sigma_t)
        +
        \beta_t^2\|r_t\|^2
    \right].
    \label{eq:lamp_risk_method}
\end{equation}
Consequently, LAMP improves the one-step posterior transition whenever
\begin{equation}
    \beta_t\|r_t\|^2
    <
    2(1-\beta_t)(1-\rho_t)\mathrm{tr}(\Sigma_t).
    \label{eq:lamp_condition_method}
\end{equation}
\end{proposition}
The condition in \eqref{eq:lamp_condition_method} captures the trade-off introduced by temporal filtering. The term $(1-\rho_t)\mathrm{tr}(\Sigma_t)$ measures the non-shared part of the local estimation error between adjacent corrected estimates, which is the component reduced by averaging. The term $\|r_t\|^2$ measures the drift of the underlying posterior target, which is the bias introduced by the lag. Thus, LAMP is beneficial when adjacent posterior targets evolve smoothly while the estimation errors are not perfectly correlated. This is the regime expected after the earliest reverse steps, and in practice we therefore apply LAMP after a short warm-up. The implementation only requires storing the previous corrected estimate and replacing $\D_t$ by the convex combination \eqref{eq:lamp_filtered_target_method}, which makes the method modular and computationally negligible.
\section{Numerical Experiments}
\label{sec:experiments}
We apply LAMP on top of DiffPIR and DDRM solvers, denoting the resulting schemes as DiffPIR-LAMP and DDRM-LAMP, and evaluate them on standard linear inverse problems in image restoration. We focus on these methods as two representative PS approaches: DiffPIR follows a plug-and-play optimization-based scheme, while DDRM implements a direct posterior sampling formulation. This choice allows us to assess the effect of LAMP across different implementations of data-consistency corrections, without requiring exhaustive evaluation over all existing posterior sampling methods.
The goal of the experiments is to test whether the proposed method improves existing diffusion posterior samplers without changing the pretrained diffusion model, the forward operator, or the number of denoising evaluations.
\subsection{Experimental setup}
\label{subsec:experimental_setup}
We compare DiffPIR-LAMP and DDRM-LAMP against their respective backbones, their first- and second-order variants (DiffPIR-1M, DiffPIR-2M, DDRM-1M, and DDRM-2M), as well as DPS. DPS is used as a general posterior sampling baseline with $1000$ denoising evaluations. DiffPIR and DDRM, together with all their variants (1M, 2M, and LAMP), are evaluated with $100$ and $20$ denoising evaluations, respectively. Since the number of function evaluations is kept fixed, any improvement achieved by LAMP comes at no additional computational cost.
We follow the experimental protocols of DiffPIR and DDRM, using their test sets and pretrained backbones.
We consider three image datasets: FFHQ, ImageNet, and CelebA. For each dataset, we evaluate three linear inverse problems: Gaussian deblurring, motion deblurring, and super-resolution with scale factor $\times4$. The measurements are generated according to $\y = \K\x_0+\boldsymbol{e}$, $\boldsymbol{e}\sim\mathcal{N}(0,\sigma_{\y}^2\I)$ with $\sigma_{\y} = 0.05$. The exact degradation parameters, preprocessing details, and hyperparameters are reported in Appendix~\ref{app:implementation}.
Additional quantitative and qualitative results are provided in Appendices \ref{app:additional_experiments} and \ref{app:qualitative}, respectively.
Precisely, Appendix~\ref{app:noiseless_experiments} reports results in the noiseless setting ($\sigma_{\y}=0$), while Appendix~\ref{app:noise_sweep} presents performance across different noise levels.
We mainly evaluate reconstruction quality using average PSNR, SSIM and LPIPS values, computed on the standard test sets. Higher PSNR and SSIM values, and lower LPIPS values, indicate better reconstruction quality. The reported values are statistically significant, as specifically validated by the results in Appendix \ref{app:robustness_std}, relative to five independent runs. In all quantitative tables, the best value in each column is shown in bold, and the second-best value is underlined.
We emphasize that although the numerical improvements in PSNR and SSIM are small, they are highly significant in this regime. Diffusion-based inverse solvers already operate near the saturation point for such metrics, and further gains are difficult to achieve. As a result, even small improvements are indicative of consistent and perceptually meaningful enhancements in reconstruction quality as can be seen in Figure ~\ref{fig:motion_noisy} and in the Appendix \ref{app:qualitative}.
All the experiments have been executed on an NVIDIA RTX 5080 GPU with 16 GB of VRAM, running PyTorch 2.11 with CUDA 13.0. The code to reproduce the experiments is available at an anonymous repository: \url{https://anonymous.4open.science/r/LAMP-Diff/}.
\subsection{Main results and discussion}
\label{subsec:main_results}
Table~\ref{tab:main_results_noisy} reports the main quantitative comparison in the noisy setting, showing that DDRM-LAMP achieves the best overall performance. While improvements in PSNR and SSIM are modest due to saturation, they are consistent across tasks. This effect is more pronounced in LPIPS, which is more sensitive to perceptual and structural differences, and better captures the gains introduced by LAMP.
In general, each LAMP version outperforms its respective backbone, as DiffPIR-LAMP yields improvements over DiffPIR in most cases, with particularly clear gains in deblurring and super-resolution, although the extent of the improvement varies depending on the dataset and degradation.
These improvements are obtained without increasing the number of denoising evaluations, highlighting the efficiency of our proposal.
The comparison with the 1M and 2M variants further clarifies this point. Naive first- and second-order ODE discretizations consistently degrade performance, indicating that simply increasing the order is not sufficient. Instead, the considered residual correction is key to achieving improved reconstructions.
\begin{table*}[h!]
\centering
\setlength{\tabcolsep}{3.5pt}
\scriptsize
\begin{tabular}{lcccccccccc}
\toprule
\multicolumn{1}{c}{\textbf{CelebA}} & \multicolumn{1}{c}{\textbf{}} & \multicolumn{3}{c}{\textbf{Deblur (Gaussian)}} & \multicolumn{3}{c}{\textbf{Deblur (Motion)}} & \multicolumn{3}{c}{\textbf{SR ($\times 4$)}} \\
\cmidrule(lr){3-5} \cmidrule(lr){6-8} \cmidrule(lr){9-11}
\textbf{Method} & NFE & PSNR$\uparrow$ & SSIM$\uparrow$ & LPIPS$\downarrow$ & PSNR$\uparrow$ & SSIM$\uparrow$ & LPIPS$\downarrow$ & PSNR$\uparrow$ & SSIM$\uparrow$ & LPIPS$\downarrow$ \\
\midrule
DPS \citep{chung2022dps} & 1000 & \textbf{26.45} & 0.720 & \textbf{0.095} & 23.74 & 0.659 & \underline{0.134} & 25.88 & 0.710 & \underline{0.118} \\
DiffPIR \citep{zhu2023diffpir} & 100 & 24.84 & 0.705 & 0.107 & 23.20 & 0.567 & 0.254 & 26.90 & 0.690 & 0.202 \\
DDRM \citep{kawar2022ddrm} & 20 & 26.11 & \underline{0.789} & 0.169 & \underline{25.52} & \underline{0.770} & 0.188 & \underline{28.97} & \underline{0.818} & 0.131 \\
DiffPIR-1M & 100 & 24.84 & 0.705 & 0.107 & 23.20 & 0.567 & 0.254 & 26.90 & 0.690 & 0.202 \\
DiffPIR-2M & 100 & 24.56 & 0.687 & 0.114 & 23.06 & 0.559 & 0.258 & 26.59 & 0.675 & 0.215 \\
DDRM-1M & 20 & 15.51 & 0.146 & 0.944 & 14.37 & 0.108 & 1.024 & 27.92 & 0.794 & 0.201 \\
DDRM-2M & 20 & 15.61 & 0.149 & 0.943 & 14.51 & 0.111 & 1.023 & 27.76 & 0.789 & 0.201 \\
\noalign{\vskip 2pt}
\cdashline{1-11}
\noalign{\vskip 2pt}
DiffPIR-LAMP (Ours) & 100 & 25.04 & 0.716 & \underline{0.106} & 23.26 & 0.566 & 0.230 & 27.11 & 0.699 & 0.191\\
DDRM-LAMP (Ours) & 20 & \underline{26.14} & \textbf{0.791} & 0.118 & \textbf{25.55} & \textbf{0.772} & \textbf{0.129} & \textbf{29.12} & \textbf{0.823} & \textbf{0.105} \\
\midrule
\multicolumn{1}{c}{\textbf{FFHQ}} & \multicolumn{1}{c}{\textbf{}} & \multicolumn{3}{c}{\textbf{Deblur (Gaussian)}} & \multicolumn{3}{c}{\textbf{Deblur (Motion)}} & \multicolumn{3}{c}{\textbf{SR ($\times 4$)}} \\
\cmidrule(lr){3-5} \cmidrule(lr){6-8} \cmidrule(lr){9-11}
\textbf{Method} & NFE & PSNR$\uparrow$ & SSIM$\uparrow$ & LPIPS$\downarrow$ & PSNR$\uparrow$ & SSIM$\uparrow$ & LPIPS$\downarrow$ & PSNR$\uparrow$ & SSIM$\uparrow$ & LPIPS$\downarrow$ \\
\midrule
DPS \citep{chung2022dps} & 1000 & 25.35 & 0.706 & \underline{0.142} & 21.58 & 0.595 & 0.204 & 24.27 & 0.676 & \underline{0.186}\\
DiffPIR \citep{zhu2023diffpir} & 100 & 24.84 & 0.741 & \textbf{0.132} & 23.66 & 0.696 & \textbf{0.170} & 26.54 & 0.676 & 0.239 \\
DDRM \citep{kawar2022ddrm} & 20 & \underline{25.51} & \underline{0.776} & 0.238 & \underline{24.48} & \underline{0.732} & 0.249 & \underline{27.94} & \underline{0.804} & 0.189 \\
DiffPIR-1M & 100 & 23.71 & 0.610 & 0.188 & 21.23 & 0.386 & 0.391 & 25.01 & 0.592 & 0.309 \\
DiffPIR-2M & 100 & 23.35 & 0.585 & 0.205 & 21.00 & 0.376 & 0.401 & 24.65 & 0.574 & 0.315 \\
DDRM-1M & 20 & 11.80 & 0.073 & 1.203 & 11.51 & 0.061 & 1.141 & 26.11 & 0.696 & 0.273 \\
DDRM-2M & 20 & 11.85 & 0.074 & 1.210 & 11.58 & 0.062 & 1.142 & 25.99 & 0.693 & 0.271 \\
\noalign{\vskip 2pt}
\cdashline{1-11}
\noalign{\vskip 2pt}
DiffPIR-LAMP (Ours) & 100 & 25.18 & 0.761 & 0.160 & 23.94 & 0.713 & 0.202 & 27.05 & 0.704 & 0.242 \\
DDRM-LAMP (Ours) & 20 & \textbf{25.55} & \textbf{0.779} & 0.164 & \textbf{24.52} & \textbf{0.737} & \underline{0.181} & \textbf{28.07} & \textbf{0.808} & \textbf{0.146} \\
\midrule
\multicolumn{1}{c}{\textbf{ImageNet}} & \multicolumn{1}{c}{\textbf{}} & \multicolumn{3}{c}{\textbf{Deblur (Gaussian)}} & \multicolumn{3}{c}{\textbf{Deblur (Motion)}} & \multicolumn{3}{c}{\textbf{SR ($\times 4$)}} \\
\cmidrule(lr){3-5} \cmidrule(lr){6-8} \cmidrule(lr){9-11}
\textbf{Method} & NFE & PSNR$\uparrow$ & SSIM$\uparrow$ & LPIPS$\downarrow$ & PSNR$\uparrow$ & SSIM$\uparrow$ & LPIPS$\downarrow$ & PSNR$\uparrow$ & SSIM$\uparrow$ & LPIPS$\downarrow$ \\
\midrule
DPS \citep{chung2022dps} & 1000 & 21.02 & 0.495 & \textbf{0.146} & 16.97 & 0.367 & 0.473 & 20.76 & 0.477 & 0.461 \\
DiffPIR \citep{zhu2023diffpir} & 100 & 21.77 & 0.581 & \underline{0.337} & 20.96 & 0.528 & \underline{0.460} & 23.17 & 0.569 & 0.481 \\
DDRM \citep{kawar2022ddrm} & 20 & \underline{22.38} & \underline{0.611} & 0.500 & \underline{21.36} & \underline{0.544} & 0.480 & \underline{23.73} & \underline{0.642} & 0.405 \\
DiffPIR-1M & 100 & 19.27 & 0.361 & 0.435 & 13.92 & 0.106 & 0.997 & 20.85 & 0.439 & 0.432 \\
DiffPIR-2M & 100 & 18.73 & 0.340 & 0.450 & 13.46 & 0.100 & 1.000 & 20.02 & 0.392 & 0.474 \\
DDRM-1M & 20 & 8.53 & 0.043 & 1.113 & 8.71 & 0.039 & 1.160 & 21.86 & 0.548 & \underline{0.401} \\
DDRM-2M & 20 & 8.50 & 0.043 & 1.115 & 8.68 & 0.039 & 1.163 & 21.70 & 0.539 & 0.405 \\
\noalign{\vskip 2pt}
\cdashline{1-11}
\noalign{\vskip 2pt}
DiffPIR-LAMP (Ours) & 100 & 22.01 & 0.596 & 0.397 & 21.01 & 0.529 & 0.485 & 23.43 & 0.588 & 0.502 \\
DDRM-LAMP (Ours) & 20 & \textbf{22.41} & \textbf{0.613} & 0.416 & \textbf{21.38} & \textbf{0.547} & \textbf{0.410} & \textbf{23.80} & \textbf{0.647} & \textbf{0.372}  \\
\bottomrule
\end{tabular}
\caption{Quantitative comparison in the noisy setting. Best values are in bold and second-best values are underlined. Errors over this measures can be found in the Table \ref{tab:appendix_robustness_noisy}. }
\label{tab:main_results_noisy}
\end{table*}
\begin{figure}[h!]
    \setlength{\tabcolsep}{2pt}
    \centering
    \scriptsize
    \begin{tabular}{@{}ccccccc@{}}
        Measured & DPS & DiffPIR & DDRM & DiffPIR-LAMP & DDRM-LAMP & Original \\
        \includegraphics[width=0.12\linewidth]{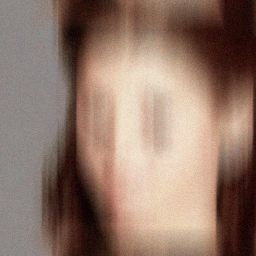} &
        \includegraphics[width=0.12\linewidth]{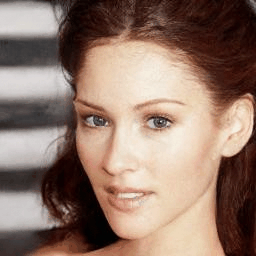} &
        \includegraphics[width=0.12\linewidth]{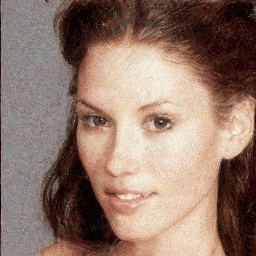} &
        \includegraphics[width=0.12\linewidth]{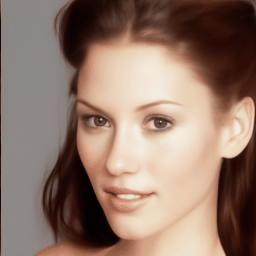} &
        \includegraphics[width=0.12\linewidth]{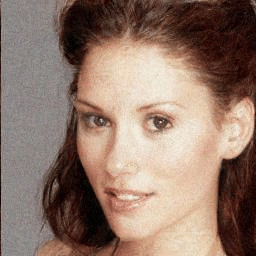} &
        \includegraphics[width=0.12\linewidth]{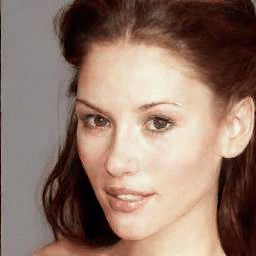} &
        \includegraphics[width=0.12\linewidth]{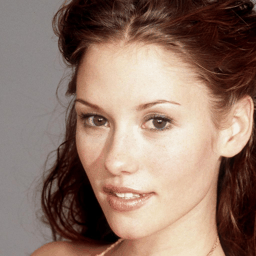} \\
        \includegraphics[width=0.12\linewidth]{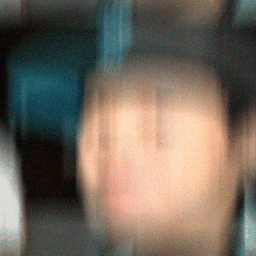} &
        \includegraphics[width=0.12\linewidth]{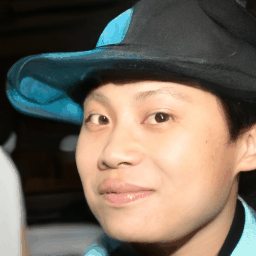} &
        \includegraphics[width=0.12\linewidth]{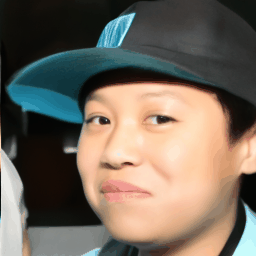} &
        \includegraphics[width=0.12\linewidth]{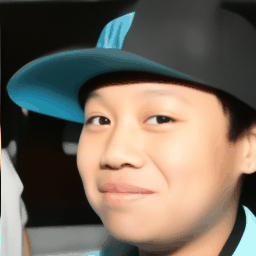} &
        \includegraphics[width=0.12\linewidth]{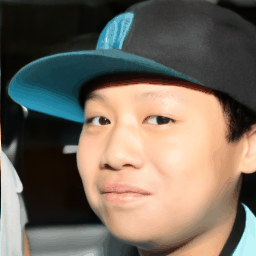} &
        \includegraphics[width=0.12\linewidth]{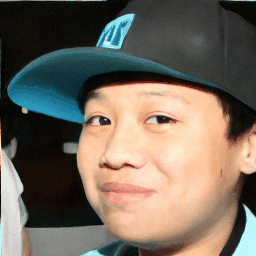} &
        \includegraphics[width=0.12\linewidth]{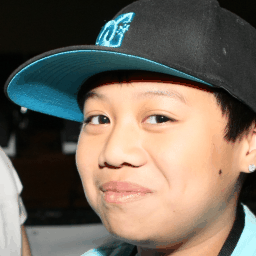} \\
        \includegraphics[width=0.12\linewidth]{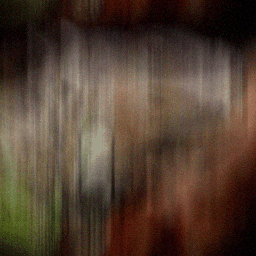} &
        \includegraphics[width=0.12\linewidth]{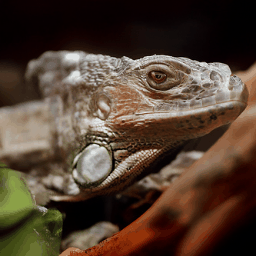} &
        \includegraphics[width=0.12\linewidth]{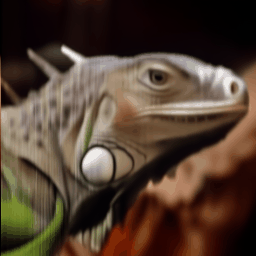} &
        \includegraphics[width=0.12\linewidth]{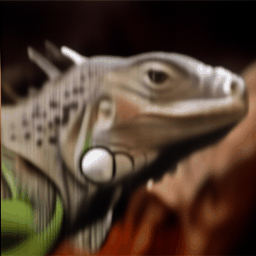} &
        \includegraphics[width=0.12\linewidth]{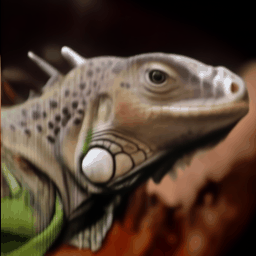} &
        \includegraphics[width=0.12\linewidth]{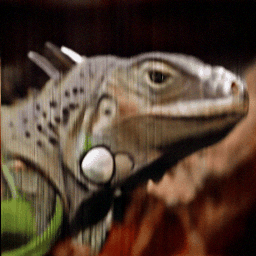} &
        \includegraphics[width=0.12\linewidth]{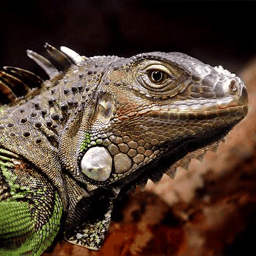} \\
    \end{tabular}
    \caption{Qualitative comparison for noisy motion deblurring across different datasets.}
    \label{fig:motion_noisy}
\end{figure}
Figure~\ref{fig:motion_noisy} shows some qualitative results for noisy motion deblurring, a particularly challenging task where temporal fluctuations in the correction sequence can significantly affect reconstruction quality.
The restored images highlight consistent improvements across different structures brought by LAMP. In the first row, DDRM-LAMP better preserves hair and produces a more natural, uniform skin texture than the smoother DDRM output, whereas DiffPIR-LAMP reduces the graininess visible in DiffPIR. In the second row, the faces have reduced blurring, and the caps exhibit sharper boundaries and more coherent shading with LAMP.
In the third row, the iguana’s texture is more clearly recovered: DiffPIR-LAMP enhances the contrast and definition of its scales, while DDRM-LAMP yields sharper dark details along the back of the neck.
Overall, both quantitative and qualitative results support the main claim of the paper: temporal filtering of the measurement-aware estimate improves posterior sampling. The gains are most consistent for DDRM-LAMP, which improves DDRM across most deblurring and super-resolution settings, while DiffPIR-LAMP yields smaller but still consistent improvements. This suggests that the benefit of LAMP depends on the temporal behavior of the correction sequence of the underlying sampler. Additional results in the appendices further support these observations.
\subsection{Ablation on the lagged correction}
\label{subsec:ablation_lagged_correction}
\begin{figure}[bthp]
    \centering
    \boxed{\includegraphics[height=0.17\textheight]{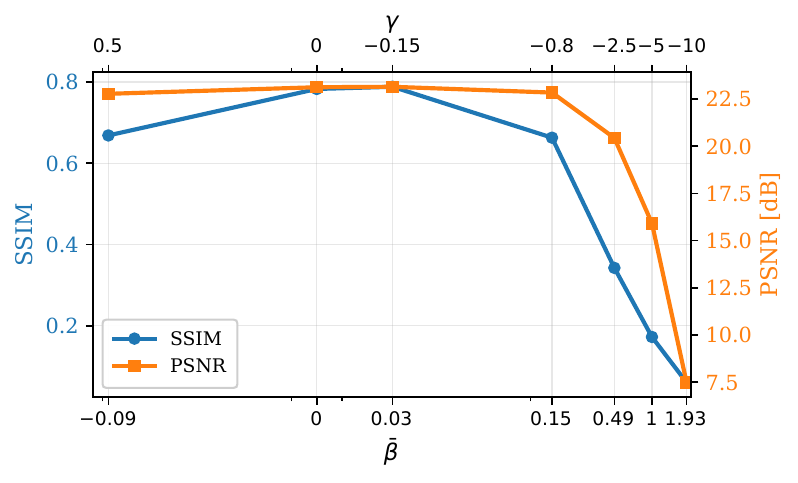}}
    \quad \boxed{\includegraphics[height=0.17\textheight]{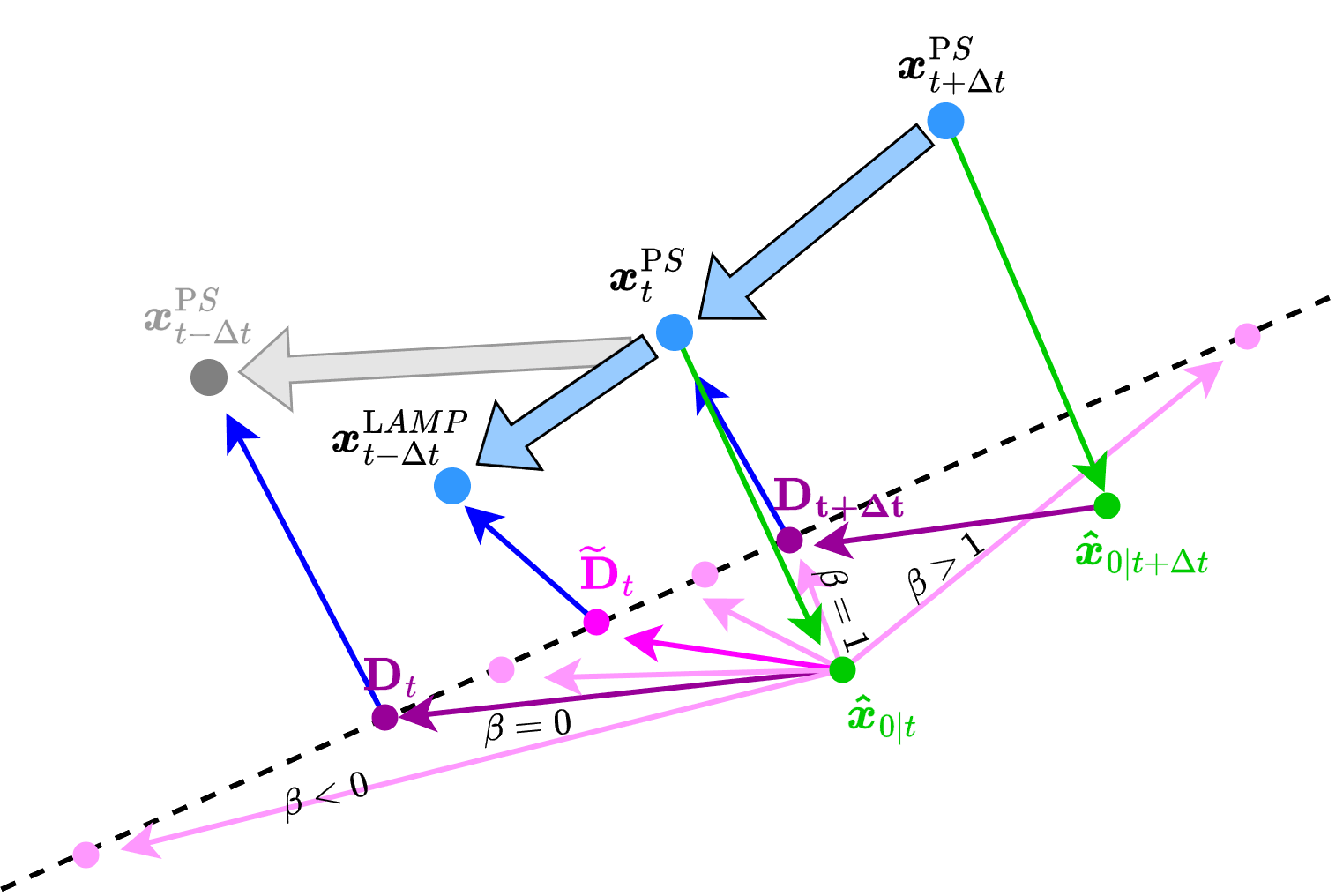}    }
    \caption{Ablation of the lagged correction strength.
Left: PSNR and SSIM as functions of the average coefficient $\bar{\beta}$ (with corresponding $\gamma$ values shown on top).
Right: geometric interpretation of the temporal correction, highlighting the role of the coefficient $\beta$ as in \eqref{eq:lamp_filtered_target_method} at each temporal update.}
    \label{fig:beta_gammaAblation}
\end{figure}

While LAMP is designed around a \emph{lagged} correction (i.e., $\gamma<0$ in \eqref{eq:lamp_from_PS_method}), the formulation itself does not restrict $\gamma$ to negative values. We therefore study its effect through an ablation analysis. Figure~\ref{fig:beta_gammaAblation} (left) reports PSNR and SSIM as functions of the correction strength, with $\gamma$ shown on the top axis, for DDRM-LAMP in the case of motion deblur with $\sigma_{\y} = 0.05$, averaged over FFHQ dataset.
Recalling the effective coefficient $\beta_t := -\gamma \frac{A_1(h)}{h_{\mathrm{prev}}}$ introduced in Proposition~\ref{prop:lamp_lagged_form}, which makes explicit the relation between $\widetilde{\D}_t$, $\D_t$, and $\D_{t+\Delta t}$, we additionally report on the bottom axis its average value $\bar{\beta}$ across iterations.
In Figure~\ref{fig:beta_gammaAblation} (right), the role of $\beta_t$ upon each update is also illustrated schematically.
As visible, the best performance is obtained with a mild lag, achieved by choosing $\gamma \approx -0.15$, matching $\gamma$ of order $\mathcal{O}(1/h)$.
This yields $\bar{\beta}\approx 0.03$, which corresponds to non-trivial yet controlled values of $\beta_t$ at each time step, within the range $(0,1)$ of moderate lagged corrections.
Specifically, if $\bar{\beta}\approx 1$, the updates exhibit a near-stalling behavior (i.e. $\tilde{\D}_t \approx \D_{t + \Delta t} \approx \tilde{\D}_{t + \Delta t}$), with the trajectory lingering around the last posterior iterate $\x^{\mathrm{PS}}_{t}$ and the final solutions degrade.
For $\bar{\beta}>1$ (i.e., large negative $\gamma$), the method introduces excessive lag, further ruining performance.
Conversely, when $\bar{\beta}=0$ (i.e. $\gamma=0$), the LAMP rule reduces to the baseline PS update and the solutions are those previously discussed.
At last, for $\bar{\beta}<0$ (i.e., $\gamma>0$), the update becomes extrapolating rather than lagged, which also degrades performance.
These results confirm that a moderate lag provides the best trade-off, supporting the design choice underlying LAMP.
\section{Limitations}
\label{sec:limitations}
LAMP is a plug-in extension of posterior sampling methods and therefore inherits their assumptions and limitations. In particular, it relies on the same modeling assumptions on the forward operator, noise distribution, and diffusion prior. As a result, it cannot compensate for failures of the underlying sampler, such as lack of convergence or model mismatch.
The effectiveness of LAMP also depends on the behavior of the backbone method. While it consistently improves performance, it does not fundamentally alter the reconstruction pattern. For example, DiffPIR-LAMP tends to preserve some of the characteristic noise patterns of DiffPIR, indicating that LAMP refines rather than replaces the underlying dynamics. Consequently, LAMP is most beneficial when the baseline sampler already produces reasonably stable estimates.
Finally, LAMP introduces a small number of additional parameters, such as the lag strength and warm-up length. Although the computational overhead is negligible and no extra denoiser evaluations are required, these parameters may require tuning when changing the sampler or the degradation setting.
\section{Conclusion}
\label{sec:conclusion}
We introduced LAMP, a lagged multistep correction for diffusion posterior samplers in image reconstruction problems. The method is motivated by the observation that standard posterior updates combine a first-order discretization with a residual forcing term, which can lead to unstable corrections when extended naively to multistep schemes.
LAMP addresses this issue, and the resulting update can still be interpreted as a standard posterior sampler, applied to a filtered estimate.
Theoretical analysis characterizes a bias-variance trade-off, showing that LAMP improves the reverse transition when temporal smoothing reduces variability more than it introduces bias.
Empirical results support this behavior and show that LAMP consistently improves reconstruction quality, with clearer gains in data fidelity for fine-scale structures, with a soft lagged correction.
Future work includes adaptive lag selection, extensions to nonlinear settings, and a deeper analysis of temporally regularized diffusion dynamics.
\bibliographystyle{plain}
\bibliography{biblio}
\newpage
\appendix
\section{Definition of the Measurement-Aware Estimate for DiffPIR and DDRM}
\label{app:dt_diffpir_ddrm}
This section describes how the measurement-aware estimate $\D_t$ is instantiated for the two posterior samplers used as LAMP backbones, namely DiffPIR and DDRM. In the main paper, we use the abstract notation
\begin{equation}
    \D_t = \mathcal{C}_t(\hat{\x}_{0|t},\y,\K),
\end{equation}
where $\hat{\x}_{0|t}$ is the denoised estimate produced by the diffusion model at reverse time $t$, and $\mathcal{C}_t$ is the data-consistency correction associated with the chosen sampler. The purpose of this section is to make this definition explicit.
\subsection{DiffPIR correction}
\label{app:diffpir_dt}
DiffPIR can be interpreted as a plug-and-play diffusion method based on half-quadratic splitting. At each reverse step, the diffusion model produces the clean-image estimate
\begin{equation}
    \hat{\x}_{0|t}
    =
    \frac{\x_t-\sigma_t\eps_\theta(\x_t,t)}{\alpha_t}.
\end{equation}
DiffPIR then computes a data-consistent reconstruction by solving a quadratic proximal problem that balances proximity to $\hat{\x}_{0|t}$ and consistency with the measurement model
\begin{equation}
    \y = \K\x_0+\boldsymbol{e}.
\end{equation}
In our notation, the DiffPIR measurement-aware estimate is
\begin{equation}
    \D_t^{\mathrm{DiffPIR}}
    =
    \arg\min_{\x}
    \frac{1}{2\sigma_{\y}^2}
    \|\y-\K\x\|_2^2
    +
    \frac{1}{2\tau_t^2}
    \|\x-\hat{\x}_{0|t}\|_2^2,
    \label{eq:app_diffpir_prox}
\end{equation}
where $\tau_t>0$ is the step-dependent regularization parameter induced by the DiffPIR schedule. Equivalently, defining $\mu_t := \frac{\sigma_{\y}^2}{\tau_t^2}$, the solution of \eqref{eq:app_diffpir_prox} can be written as
\begin{align}
    \D_t^{\mathrm{DiffPIR}}
    &=
    \arg\min_{\x}
    \frac{1}{2}
    \|\y-\K\x\|_2^2
    +
    \frac{\mu_t}{2}
    \|\x-\hat{\x}_{0|t}\|_2^2 \\
    &=
    \left(
        \K^T\K+\mu_t\I
    \right)^{-1}
    \left(
        \K^T\y+\mu_t\hat{\x}_{0|t}
    \right).
    \label{eq:app_diffpir_closed_form}
\end{align}
Thus, in DiffPIR, $\D_t$ is the proximal reconstruction obtained by pulling the denoised estimate toward the affine measurement constraint.
The exact implementation of \eqref{eq:app_diffpir_closed_form} depends on the forward operator. For convolutional blur, the linear system is solved efficiently in the Fourier domain; for super-resolution, one can exploit the operator structure to solve it exactly.
\subsection{DDRM spectral correction and operator decompositions}
\label{app:ddrm}
DDRM \citep{kawar2022ddrm} is a posterior sampling method for linear inverse problems of the form
\begin{equation}
    \y = \K \x_0 + \boldsymbol{e},
    \qquad
    \boldsymbol{e}\sim\mathcal{N}(0,n_0^2\I),
\end{equation}
where $n_0=\sigma_{\y}$ denotes the observation noise level. Its key idea is to exploit a spectral decomposition of the linear degradation operator, so that the inverse problem decouples into one-dimensional problems. Let
\begin{equation}
    \K = U\boldsymbol{\Sigma}V^T
\end{equation}
be a singular value decomposition of $\K$, where $\boldsymbol{\Sigma}$ contains the singular values $\{a_i\}_{i=1}^r$. In the corresponding coordinates,
\begin{equation}
    \bar{\y}:=U^T\y,
    \qquad
    \bar{\x}_{0|t}:=V^T\hat{\x}_{0|t},
\end{equation}
the measurement model becomes componentwise
\begin{equation}
    \bar{y}_i=a_i\bar{x}_{0,i}+\bar{e}_i.
\end{equation}
Thus, each spectral component can be corrected independently according to the singular value $a_i$ and the relative strength of the diffusion and measurement noise.
In the variance-preserving parameterization used in this paper, the effective diffusion noise level at time $t$ is
\begin{equation}
    n_t:=\frac{\sigma_t}{\alpha_t}.
\end{equation}
DDRM combines the denoiser prediction $\hat{\x}_{0|t}$ with the measurement in the $V$-coordinates. Writing $\bar{\D}_{t,i}^{\mathrm{DDRM}} := \left(V^T\D_t^{\mathrm{DDRM}}\right)_i$, the deterministic measurement-aware estimate used by the posterior update is defined componentwise as follows:
\begin{itemize}
    \item For components that are not observed by the forward operator, namely when $a_i=0$, the measurement provides no information and DDRM keeps the prior estimate:
        \begin{equation}
            \bar{\D}_{t,i}^{\mathrm{DDRM}}
            =
            \bar{x}_{0|t,i}.
            \label{eq:app_ddrm_unconstrained}
    \end{equation}
    \item For observed components satisfying
        \begin{equation}
            a_i n_t > n_0,
        \end{equation}
        the current diffusion uncertainty is large enough compared with the observation noise that DDRM can safely move the estimate toward the pseudo-inverse measurement. In this regime,
        \begin{equation}
            \bar{\D}_{t,i}^{\mathrm{DDRM}}
            =
            \eta_B\frac{\bar{y}_i}{a_i}
            +
            (1-\eta_B)\bar{x}_{0|t,i},
            \label{eq:app_ddrm_pinv}
        \end{equation}
        where $\eta_B\in[0,1]$ controls the strength of the data-consistency replacement. The choice $\eta_B=1$ gives the fully data-consistent value $\bar{y}_i/a_i$.
    \item For observed components satisfying
        \begin{equation}
            a_i n_t \leq n_0,
        \end{equation}
        the measurement noise is large relative to the current diffusion level along that spectral direction. DDRM therefore uses a residual correction rather than a direct pseudo-inverse replacement:
        \begin{equation}
            \bar{\D}_{t,i}^{\mathrm{DDRM}}
            =
            \bar{x}_{0|t,i}
            +
            \widetilde{n}_t
            \frac{\bar{y}_i-a_i\bar{x}_{0|t,i}}{n_0},
            \label{eq:app_ddrm_residual}
        \end{equation}
        with
        \begin{equation}
            \widetilde{n}_t
            =
            n_t\sqrt{1-\eta^2}.
            \label{eq:app_ddrm_tilde_noise}
        \end{equation}
        Here $\eta\in[0,1]$ is the stochasticity parameter of DDRM. The complete DDRM sampler may additionally inject stochastic terms in the spectral coordinates, while the quantity $\D_t^{\mathrm{DDRM}}$ above is the deterministic measurement-aware target that enters the posterior update.
\end{itemize}
After applying the componentwise correction, the corrected estimate is mapped back to the image domain through
\begin{equation}
    \D_t^{\mathrm{DDRM}}
    =
    V\bar{\D}_t^{\mathrm{DDRM}}.
    \label{eq:app_ddrm_D}
\end{equation}
This construction shows that DDRM balances prior information and data consistency in a spectral way: directions that are well constrained by the measurements are corrected toward the pseudo-inverse solution, while directions that are weakly observed, noisy, or in the null space remain closer to the diffusion prior.
One crucial limitation of DDRM is that it requires access to the SVD decomposition of the operator $\K$. For the inverse problems considered in this paper, the required spectral structure is available in closed form or through standard structured decompositions.
In particular, for Gaussian and motion deblurring, the degradation operator is a convolution. Under circular boundary conditions, convolution is diagonalized by the two-dimensional discrete Fourier transform. If $k$ is the blur kernel padded to the image size and
\begin{equation}
    \widehat{k}=\mathcal{F}(k),
\end{equation}
then
\begin{equation}
    \mathcal{F}(\K\x)(\omega)
    =
    \widehat{k}(\omega)\mathcal{F}(\x)(\omega).
\end{equation}
Writing
\begin{equation}
    \widehat{k}(\omega)
    =
    a(\omega)e^{\mathrm{i}\phi(\omega)},
    \qquad
    a(\omega)=|\widehat{k}(\omega)|,
\end{equation}
the singular values are given by the Fourier magnitudes $a(\omega)$, while the phase factors are absorbed into the unitary matrices of the SVD. Equivalently, after rotating the Fourier-domain measurements by the conjugate phase, each frequency obeys a scalar degradation model with singular value $a(\omega)$. DDRM then applies the same componentwise correction rules in \eqref{eq:app_ddrm_unconstrained}, \eqref{eq:app_ddrm_pinv}, and \eqref{eq:app_ddrm_residual} frequency by frequency, and the corrected estimate is transformed back to the image domain by the inverse Fourier transform. This Fourier representation is exact for circular convolution. Therefore, the DDRM formulation for deblurring should be understood under the circular-convolution model associated with the padded blur kernel. With other boundary conditions, such as zero or reflection padding, the Fourier diagonalization is no longer exact, and the same construction becomes an approximation unless the corresponding operator is explicitly diagonalized or otherwise adapted.
For super-resolution by a factor $r$, the forward operator averages each non-overlapping $r\times r$ block and then downsamples. The local degradation acting on one block is the rank-one operator
\begin{equation}
    H_{\mathrm{SR}}
    =
    \begin{bmatrix}
        \frac{1}{r^2} & \frac{1}{r^2} & \cdots & \frac{1}{r^2}
    \end{bmatrix}
    \in\mathbb{R}^{1\times r^2}.
\end{equation}
This operator admits a simple blockwise SVD: one direction, corresponding to the average intensity in the block, has a nonzero singular value, while the remaining $r^2-1$ directions span the null space. Applying this decomposition independently across blocks and channels yields a structured spectral representation of the full super-resolution operator. DDRM therefore corrects the block-average component using the measurement, while the high-frequency within-block components are sampled from the diffusion prior.
In our framework, LAMP does not modify the DDRM correction itself. Instead, it takes the DDRM measurement-aware target $\D_t^{\mathrm{DDRM}}$ and replaces it by the lagged estimate
\begin{equation}
    \widetilde{\D}_t^{\mathrm{DDRM}}
    =
    (1-\beta_t)\D_t^{\mathrm{DDRM}}
    +
    \beta_t\D_{t+\Delta t}^{\mathrm{DDRM}},
\end{equation}
which is then inserted into the same posterior sampling update:
\begin{equation}
    \x_{t-\Delta t}^{\mathrm{DDRM-LAMP}}
    =
    \alpha_{t-\Delta t}\widetilde{\D}_t^{\mathrm{DDRM}}
    +
    \sigma_{t-\Delta t}\eps_\Theta(\x_t,t).
\end{equation}
Thus, DDRM-LAMP remains a DDRM-type posterior sampler, but with the instantaneous spectral correction replaced by a temporally filtered one.
\section{Proofs}
\label{app:proofs}
This section provides the full derivations of the main structural identities used in the methodological section \ref{sec:methodology}. In the following, we use the same notation as in the main paper. In particular, $t$ and $t-\Delta t$ denote two consecutive reverse diffusion times, $h=\lambda_{t-\Delta t}-\lambda_t>0$, and
\[
    \lambda_t=\log\!\left(\frac{\alpha_t}{\sigma_t}\right).
\]
\subsection{\texorpdfstring{Proof of Eq.~\eqref{eq:mild_ddim_method}}{Proof of Eq. (mild DDIM method)}}
\label{app:proof_if_form}
While this result is not new, we report for completeness the derivation of the integrating-factor formulation. For a variance-preserving diffusion, the probability-flow ODE can be written in the clean-image parameterization as
\begin{equation}
    \frac{d\x}{dt}
    =
    -\frac{1}{2}\beta(t)\x
    +
    \frac{1}{2}\beta(t)
    \frac{\x-\alpha_t\hat{\x}_0(\x,t)}{\sigma_t^2}.
    \label{eq:app_pfode_x0}
\end{equation}
Using the variance-preserving identities
\begin{equation}
    \frac{d\alpha_t}{dt}
    =
    -\frac{1}{2}\beta(t)\alpha_t,
    \qquad
    \frac{d\sigma_t^2}{dt}
    =
    \beta(t)\alpha_t^2,
\end{equation}
one obtains
\begin{equation}
    \frac{d\lambda_t}{dt}
    =
    -\frac{1}{2}\beta(t)
    \left(
        1+\frac{\alpha_t^2}{\sigma_t^2}
    \right)
    =
    -\frac{\beta(t)}{2\sigma_t^2},
    \label{eq:app_lambda_derivative}
\end{equation}
where the last equality uses $\alpha_t^2+\sigma_t^2=1$. Reparametrizing \eqref{eq:app_pfode_x0} by $\lambda$ gives
\begin{equation}
    \frac{d\x}{d\lambda}
    =
    \frac{d\x/dt}{d\lambda/dt}
    =
    \x-\alpha_\lambda\hat{\x}_0(\x,\lambda)
    -
    \sigma_\lambda^2\x.
\end{equation}
Since $1-\sigma_\lambda^2=\alpha_\lambda^2$, this becomes
\begin{equation}
    \frac{d\x}{d\lambda}
    =
    \alpha_\lambda^2\x
    -
    \alpha_\lambda\hat{\x}_0(\x,\lambda).
    \label{eq:app_dx_dlambda}
\end{equation}
Depending on the convention used to orient the reverse-time integration, the same equation is equivalently written with the integration limits ordered from the noisier time $t$ to the cleaner time $t-\Delta t$. Under the convention adopted in the main paper, this yields the integrating-factor identity
\begin{equation}
    \frac{d}{d\lambda}
    \left(
        \frac{\x(\lambda)}{\sigma(\lambda)}
    \right)
    =
    \frac{\alpha(\lambda)}{\sigma(\lambda)}
    \hat{\x}_0(\x(\lambda),\lambda).
    \label{eq:app_if_form}
\end{equation}
The mild formulation follows by integrating \eqref{eq:app_if_form}:
\begin{equation}
    \x_{t-\Delta t}
    =
    \frac{\sigma_{t-\Delta t}}{\sigma_t}\x_t
    +
    \sigma_{t-\Delta t}
    \int_{\lambda_t}^{\lambda_{t-\Delta t}}
    \frac{\alpha(\lambda)}{\sigma(\lambda)}
    \hat{\x}_0(\x(\lambda),\lambda)
    \,d\lambda .
    \label{eq:app_mild_form}
\end{equation}
Approximating $\hat{\x}_0(\x(\lambda),\lambda)$ by $\hat{\x}_{0|t}$ on the integration interval gives exactly \eqref{eq:ddim_1m_method}.
\subsection{\texorpdfstring{Proof of Eq.~\ref{eq:posterior_1m_method}}{Proof of the posterior 1M method}}
\label{app:proof_ddim_1m}
The deterministic DDIM step is
\begin{equation}
    \x_{t-\Delta t}
    =
    \alpha_{t-\Delta t}\hat{\x}_{0|t}
    +
    \sigma_{t-\Delta t}\eps_\Theta(\x_t,t),
    \label{eq:app_ddim_step}
\end{equation}
where
\begin{equation}
    \hat{\x}_{0|t}
    =
    \frac{\x_t-\sigma_t\eps_\Theta(\x_t,t)}{\alpha_t}.
    \label{eq:app_x0hat}
\end{equation}
Rearranging \eqref{eq:app_x0hat} gives
\begin{equation}
    \eps_\Theta(\x_t,t)
    =
    \frac{\x_t-\alpha_t\hat{\x}_{0|t}}{\sigma_t}.
    \label{eq:app_eps_identity}
\end{equation}
Substituting \eqref{eq:app_eps_identity} into \eqref{eq:app_ddim_step}, we obtain
\begin{align}
    \x_{t-\Delta t}
    &=
    \alpha_{t-\Delta t}\hat{\x}_{0|t}
    +
    \sigma_{t-\Delta t}
    \frac{\x_t-\alpha_t\hat{\x}_{0|t}}{\sigma_t}
    \nonumber\\
    &=
    \frac{\sigma_{t-\Delta t}}{\sigma_t}\x_t
    +
    \left(
        \alpha_{t-\Delta t}
        -
        \frac{\sigma_{t-\Delta t}\alpha_t}{\sigma_t}
    \right)
    \hat{\x}_{0|t}.
    \label{eq:app_ddim_intermediate}
\end{align}
Using the definition of the log-SNR,
\begin{equation}
    e^{-h}
    =
    e^{-(\lambda_{t-\Delta t}-\lambda_t)}
    =
    \frac{\alpha_t\sigma_{t-\Delta t}}{\alpha_{t-\Delta t}\sigma_t},
    \label{eq:app_exp_h_identity}
\end{equation}
we have
\begin{equation}
    \frac{\sigma_{t-\Delta t}\alpha_t}{\sigma_t}
    =
    \alpha_{t-\Delta t} e^{-h}.
    \label{eq:app_alpha_identity}
\end{equation}
Therefore,
\begin{equation}
    \alpha_{t-\Delta t}
    -
    \frac{\sigma_{t-\Delta t}\alpha_t}{\sigma_t}
    =
    \alpha_{t-\Delta t}(1-e^{-h})
    =
    \alpha_{t-\Delta t} A_0(h),
    \qquad
    A_0(h)=1-e^{-h}.
\end{equation}
Substituting this into \eqref{eq:app_ddim_intermediate} gives
\begin{equation}
    \x_{t-\Delta t}
    =
    \frac{\sigma_{t-\Delta t}}{\sigma_t}\x_t
    +
    \alpha_{t-\Delta t} A_0(h)\hat{\x}_{0|t}.
    \label{eq:app_ddim_1m}
\end{equation}
This is the first-order exponential-integrator form of DDIM.
\subsection{\texorpdfstring{Proof of Eq.~\eqref{eq:posterior_PS_1M_forcing_method}}{Proof of Eq. (Posterior Forcing Method)}}
\label{app:proof_posterior_forcing}
We now prove the structural identity used to motivate LAMP. The generic posterior sampler step is
\begin{equation}
    \x_{t-\Delta t}^{\mathrm{PS}}
    =
    \alpha_{t-\Delta t}\D_t
    +
    \sigma_{t-\Delta t}\eps_\Theta(\x_t,t).
    \label{eq:app_ps_step}
\end{equation}
Using again
\begin{equation}
    \eps_\Theta(\x_t,t)
    =
    \frac{\x_t-\alpha_t\hat{\x}_{0|t}}{\sigma_t},
\end{equation}
we obtain
\begin{align}
    \x_{t-\Delta t}^{\mathrm{PS}}
    &=
    \alpha_{t-\Delta t}\D_t
    +
    \frac{\sigma_{t-\Delta t}}{\sigma_t}\x_t
    -
    \frac{\sigma_{t-\Delta t}\alpha_t}{\sigma_t}
    \hat{\x}_{0|t}
    \nonumber\\
    &=
    \frac{\sigma_{t-\Delta t}}{\sigma_t}\x_t
    +
    \alpha_{t-\Delta t}\D_t
    -
    \alpha_{t-\Delta t} e^{-h}\hat{\x}_{0|t},
    \label{eq:app_ps_intermediate}
\end{align}
where we used \eqref{eq:app_alpha_identity}. We now add and subtract $\alpha_{t-\Delta t} e^{-h}\D_t$:
\begin{align}
    \x_{t-\Delta t}^{\mathrm{PS}}
    &=
    \frac{\sigma_{t-\Delta t}}{\sigma_t}\x_t
    +
    \alpha_{t-\Delta t}\D_t
    -
    \alpha_{t-\Delta t} e^{-h}\D_t
    +
    \alpha_{t-\Delta t} e^{-h}\D_t
    -
    \alpha_{t-\Delta t} e^{-h}\hat{\x}_{0|t}
    \nonumber\\
    &=
    \frac{\sigma_{t-\Delta t}}{\sigma_t}\x_t
    +
    \alpha_{t-\Delta t}(1-e^{-h})\D_t
    +
    \alpha_{t-\Delta t} e^{-h}
    \left(
        \D_t-\hat{\x}_{0|t}
    \right)
    \nonumber\\
    &=
    \frac{\sigma_{t-\Delta t}}{\sigma_t}\x_t
    +
    \alpha_{t-\Delta t} A_0(h)\D_t
    +
    \alpha_{t-\Delta t} e^{-h}
    \left(
        \D_t-\hat{\x}_{0|t}
    \right).
    \label{eq:app_posterior_forcing}
\end{align}
This proves the decomposition. The first two terms coincide with the first-order exponential step associated with the modified continuous equation driven by $\D_\lambda$, while the last term is the additional posterior forcing term.
\subsection{Proof of Proposition \ref{prop:lamp_lagged_form}}
\label{app:proof_lagged_form}
The LAMP update is defined as
\begin{equation}
    \x_{t-\Delta t}^{\mathrm{LAMP}}
    =
    \x_{t-\Delta t}^{\mathrm{PS}}
    +
    \alpha_{t-\Delta t} A_1(h)\gamma
    \frac{\D_t-\D_{t+\Delta t}}{h_{\mathrm{prev}}},
    \qquad
    \gamma<0.
    \label{eq:app_lamp_update}
\end{equation}
Using
\begin{equation}
    \x_{t-\Delta t}^{\mathrm{PS}}
    =
    \alpha_{t-\Delta t}\D_t
    +
    \sigma_{t-\Delta t}\eps_\Theta(\x_t,t),
\end{equation}
we get
\begin{align}
    \x_{t-\Delta t}^{\mathrm{LAMP}}
    &=
    \alpha_{t-\Delta t}\D_t
    +
    \sigma_{t-\Delta t}\eps_\Theta(\x_t,t)
    +
    \alpha_{t-\Delta t} A_1(h)\gamma
    \frac{\D_t-\D_{t+\Delta t}}{h_{\mathrm{prev}}}
    \nonumber\\
    &=
    \alpha_{t-\Delta t}
    \left[
        \D_t
        +
        \gamma
        \frac{A_1(h)}{h_{\mathrm{prev}}}
        (\D_t-\D_{t+\Delta t})
    \right]
    +
    \sigma_{t-\Delta t}\eps_\Theta(\x_t,t).
    \label{eq:app_lamp_intermediate}
\end{align}
Define
\begin{equation}
    \beta_t
    :=
    -\gamma
    \frac{A_1(h)}{h_{\mathrm{prev}}}.
    \label{eq:app_beta}
\end{equation}
Since $\gamma<0$ and $A_1(h)/h_{\mathrm{prev}}>0$ under the reverse schedule used in the main paper, we have $\beta_t>0$. The term in brackets in \eqref{eq:app_lamp_intermediate} becomes
\begin{equation}
    \D_t-\beta_t(\D_t-\D_{t+\Delta t})
    =
    (1-\beta_t)\D_t+\beta_t\D_{t+\Delta t}.
\end{equation}
Defining
\begin{equation}
    \widetilde{\D}_t
    :=
    (1-\beta_t)\D_t+\beta_t\D_{t+\Delta t},
\end{equation}
we obtain
\begin{equation}
    \x_{t-\Delta t}^{\mathrm{LAMP}}
    =
    \alpha_{t-\Delta t}\widetilde{\D}_t
    +
    \sigma_{t-\Delta t}\eps_\Theta(\x_t,t).
    \label{eq:app_lamp_compact}
\end{equation}
Applying the posterior forcing identity \eqref{eq:app_posterior_forcing} with $\D_t$ replaced by $\widetilde{\D}_t$ also gives
\begin{equation}
    \x_{t-\Delta t}^{\mathrm{LAMP}}
    =
    \frac{\sigma_{t-\Delta t}}{\sigma_t}\x_t
    +
    \alpha_{t-\Delta t} A_0(h)\widetilde{\D}_t
    +
    \alpha_{t-\Delta t} e^{-h}
    \left(
        \widetilde{\D}_t-\hat{\x}_{0|t}
    \right).
    \label{eq:app_lamp_forcing}
\end{equation}
This proves that LAMP preserves the posterior sampling structure.
\subsection{Variance reduction under the local stationarity model}
\label{app:proof_variance_reduction}
In this section we prove that, under the same assumptions of Proposition \ref{prop:lamp_risk_method}, the LAMP-modified correction term $\widetilde{\D}_t = (1 - \beta)\D_t + \beta \D_{t+\Delta t}$ has less variance then $\D_t$. We consider the local error model
\begin{equation}
    \D_t=\mu_t+\eta_t,
    \qquad
    \D_{t+\Delta t}=\mu_{t+\Delta t}+\eta_{t+\Delta t},
\end{equation}
with
\begin{equation}
    \E[\eta_t]=\E[\eta_{t+\Delta t}]=0,
    \qquad
    \mathrm{Var}(\eta_t)=\mathrm{Var}(\eta_{t+\Delta t})=\Sigma_t,
    \qquad
    \mathrm{Cov}(\eta_t,\eta_{t+\Delta t})=\rho_t\Sigma_t.
\end{equation}
The lagged estimate is
\begin{equation}
    \widetilde{\D}_t
    =
    (1-\beta_t)\D_t+\beta_t\D_{t+\Delta t}.
\end{equation}
The fluctuation part of $\widetilde{\D}_t$ is
\begin{equation}
    \widetilde{\eta}_t
    =
    (1-\beta_t)\eta_t+\beta_t\eta_{t+\Delta t}.
\end{equation}
Therefore,
\begin{align}
    \mathrm{Var}(\widetilde{\eta}_t)
    &=
    (1-\beta_t)^2\Sigma_t
    +
    \beta_t^2\Sigma_t
    +
    2\beta_t(1-\beta_t)\rho_t\Sigma_t
    \nonumber\\
    &=
    \left[
        (1-\beta_t)^2+\beta_t^2+2\beta_t(1-\beta_t)\rho_t
    \right]\Sigma_t
    \nonumber\\
    &=
    \left[
        1-2\beta_t(1-\beta_t)(1-\rho_t)
    \right]\Sigma_t.
    \label{eq:app_variance_reduction}
\end{align}
If $0<\beta_t<1$ and $\rho_t<1$, then
\begin{equation}
    1-2\beta_t(1-\beta_t)(1-\rho_t)<1,
\end{equation}
so the lagged estimate has smaller local error variance than the instantaneous estimate.
\subsection{Proof of Proposition \ref{prop:lamp_risk_method}}
\label{app:proof_risk_comparison}
Let
\begin{equation}
    \x_{t-\Delta t}^\mu
    :=
    \alpha_{t-\Delta t}\mu_t+\sigma_{t-\Delta t}\eps_\Theta(\x_t,t)
\end{equation}
be the ideal one-step target associated with the local posterior target $\mu_t$. The base posterior step is
\begin{equation}
    \x_{t-\Delta t}^{\mathrm{PS}}
    =
    \alpha_{t-\Delta t}\D_t+\sigma_{t-\Delta t}\eps_\Theta(\x_t,t).
\end{equation}
Thus,
\begin{equation}
    \x_{t-\Delta t}^{\mathrm{PS}}-\x_{t-\Delta t}^\mu
    =
    \alpha_{t-\Delta t}(\D_t-\mu_t)
    =
    \alpha_{t-\Delta t}\eta_t.
\end{equation}
Taking expectations gives
\begin{equation}
    \E\|\x_{t-\Delta t}^{\mathrm{PS}}-\x_{t-\Delta t}^\mu\|^2
    =
    \alpha_{t-\Delta t}^2\E\|\eta_t\|^2
    =
    \alpha_{t-\Delta t}^2\mathrm{tr}(\Sigma_t).
    \label{eq:app_base_risk}
\end{equation}
For LAMP, we have
\begin{equation}
    \x_{t-\Delta t}^{\mathrm{LAMP}}
    =
    \alpha_{t-\Delta t}\widetilde{\D}_t
    +
    \sigma_{t-\Delta t}\eps_\Theta(\x_t,t),
\end{equation}
where
\begin{equation}
    \widetilde{\D}_t
    =
    (1-\beta_t)\D_t+\beta_t\D_{t+\Delta t}.
\end{equation}
Let
\begin{equation}
    r_t:=\mu_{t+\Delta t}-\mu_t.
\end{equation}
Then
\begin{align}
    \widetilde{\D}_t-\mu_t
    &=
    (1-\beta_t)(\mu_t+\eta_t)
    +
    \beta_t(\mu_{t+\Delta t}+\eta_{t+\Delta t})
    -
    \mu_t
    \nonumber\\
    &=
    (1-\beta_t)\eta_t
    +
    \beta_t\eta_{t+\Delta t}
    +
    \beta_t(\mu_{t+\Delta t}-\mu_t)
    \nonumber\\
    &=
    (1-\beta_t)\eta_t
    +
    \beta_t\eta_{t+\Delta t}
    +
    \beta_t r_t.
    \label{eq:app_lamp_error}
\end{align}
Since $\eta_t$ and $\eta_{t+\Delta t}$ are zero-mean, the cross terms between the stochastic error and the deterministic drift vanish in expectation. Therefore,
\begin{align}
    \E\|\widetilde{\D}_t-\mu_t\|^2
    &=
    \E\left\|
        (1-\beta_t)\eta_t+\beta_t\eta_{t+\Delta t}
    \right\|^2
    +
    \beta_t^2\|r_t\|^2
    \nonumber\\
    &=
    \left[
        1-2\beta_t(1-\beta_t)(1-\rho_t)
    \right]\mathrm{tr}(\Sigma_t)
    +
    \beta_t^2\|r_t\|^2,
    \label{eq:app_lamp_estimator_risk}
\end{align}
where the last equality follows from \eqref{eq:app_variance_reduction}. Multiplying by $\alpha_{t-\Delta t}^2$ gives
\begin{equation}
    \E\|\x_{t-\Delta t}^{\mathrm{LAMP}}-\x_{t-\Delta t}^\mu\|^2
    =
    \alpha_{t-\Delta t}^2
    \left[
        \left(
            1-2\beta_t(1-\beta_t)(1-\rho_t)
        \right)
        \mathrm{tr}(\Sigma_t)
        +
        \beta_t^2\|r_t\|^2
    \right].
    \label{eq:app_lamp_risk}
\end{equation}
Comparing \eqref{eq:app_lamp_risk} with \eqref{eq:app_base_risk}, LAMP improves the one-step risk whenever
\begin{equation}
    \left[
        1-2\beta_t(1-\beta_t)(1-\rho_t)
    \right]
    \mathrm{tr}(\Sigma_t)
    +
    \beta_t^2\|r_t\|^2
    <
    \mathrm{tr}(\Sigma_t).
\end{equation}
After rearranging and dividing by $\beta_t>0$, this becomes
\begin{equation}
    \beta_t\|r_t\|^2
    <
    2(1-\beta_t)(1-\rho_t)\mathrm{tr}(\Sigma_t).
\end{equation}
This proves the proposition.
\subsection{Proof of generalized version of Proposition \ref{prop:lamp_risk_method}}
\label{app:proof_unequal_variances}
We now remove the local equal-variance assumption. Assume
\begin{equation}
    \D_t=\mu_t+\eta_t,
    \qquad
    \D_{t+\Delta t}=\mu_{t+\Delta t}+\eta_{t+\Delta t},
    \qquad
    r_t:=\mu_{t+\Delta t}-\mu_t,
\end{equation}
with
\begin{equation}
    \E[\eta_t]=\E[\eta_{t+\Delta t}]=0,
    \qquad
    \Sigma_t:=\mathrm{Var}(\eta_t),
    \qquad
    \Sigma_{t+\Delta t}:=\mathrm{Var}(\eta_{t+\Delta t}),
\end{equation}
and
\begin{equation}
    C_t:=\mathrm{Cov}(\eta_t,\eta_{t+\Delta t}).
\end{equation}
The lagged estimate satisfies
\begin{equation}
    \widetilde{\D}_t-\mu_t
    =
    (1-\beta_t)\eta_t
    +
    \beta_t\eta_{t+\Delta t}
    +
    \beta_t r_t.
\end{equation}
Therefore,
\begin{align}
    \E\|\widetilde{\D}_t-\mu_t\|^2
    &=
    \beta_t^2\|r_t\|^2
    +
    (1-\beta_t)^2\mathrm{tr}(\Sigma_t)
    +
    \beta_t^2\mathrm{tr}(\Sigma_{t+\Delta t})
    \nonumber\\
    &\quad
    +
    2\beta_t(1-\beta_t)\mathrm{tr}(C_t).
    \label{eq:app_unequal_risk}
\end{align}
The base estimate has risk
\begin{equation}
    \E\|\D_t-\mu_t\|^2
    =
    \mathrm{tr}(\Sigma_t).
\end{equation}
Thus, the lagged estimate improves over the instantaneous estimate whenever
\begin{align}
    &\beta_t^2\|r_t\|^2
    +
    (1-\beta_t)^2\mathrm{tr}(\Sigma_t)
    +
    \beta_t^2\mathrm{tr}(\Sigma_{t+\Delta t})
    +
    2\beta_t(1-\beta_t)\mathrm{tr}(C_t)
    \nonumber\\
    &\hspace{4cm}
    <
    \mathrm{tr}(\Sigma_t).
\end{align}
Equivalently,
\begin{equation}
    2\beta_t\mathrm{tr}(\Sigma_t)
    >
    \beta_t^2\|r_t\|^2
    +
    \beta_t^2\mathrm{tr}(\Sigma_t)
    +
    \beta_t^2\mathrm{tr}(\Sigma_{t+\Delta t})
    +
    2\beta_t(1-\beta_t)\mathrm{tr}(C_t).
    \label{eq:app_unequal_condition}
\end{equation}
For $\beta_t>0$, this can be written as
\begin{equation}
    \beta_t
    \left(
        \|r_t\|^2
        +
        \mathrm{tr}(\Sigma_t)
        +
        \mathrm{tr}(\Sigma_{t+\Delta t})
    \right)
    <
    2
    \left(
        \mathrm{tr}(\Sigma_t)
        -
        (1-\beta_t)\mathrm{tr}(C_t)
    \right).
    \label{eq:app_unequal_condition_alt}
\end{equation}
The corresponding state-level comparison is obtained by multiplying both sides of the estimator risks by $\alpha_{t-\Delta t}^2$.
\section{Implementation Details and Hyperparameters}
\label{app:implementation}
This section reports the implementation details used in the experiments.
\subsection{LAMP Update Pseudocode}
\label{app:lamp_update}
At each reverse step, the base sampler computes the measurement-aware estimate
\begin{equation}
    \D_t=\mathcal{C}_t(\hat{\x}_{0|t},\y,\K).
\end{equation}
After a warm-up phase, LAMP replaces $\D_t$ with
\begin{equation}
    \widetilde{\D}_t
    =
    (1-\beta_t)\D_t+\beta_t\D_{t+\Delta t},
\end{equation}
and then performs the posterior update
\begin{equation}
    \x_{t-\Delta t}^{\mathrm{LAMP}}
    =
    \alpha_{t-\Delta t}\widetilde{\D}_t+\sigma_{t-\Delta t}\eps_\Theta(\x_t,t).
\end{equation}
Here, $t+\Delta t$ denotes the previous reverse time, so that $\D_{t+\Delta t}$ is already available.
\begin{algorithm}[h]
\caption{LAMP posterior sampling}
\label{alg:lamp}
\begin{algorithmic}[1]
\STATE \textbf{Input:} measurement $\y$, operator $\K$, pretrained denoiser $\eps_\Theta$, timestep $\Delta t > 0$, backbone correction $\mathcal{C}_t$, lag parameter $\beta$, warm-up length $N_{\mathrm{warm}}$
\STATE Initialize $\x_{T}\sim\mathcal{N}(0,\I)$
\STATE Set $\D_{\mathrm{prev}}=\texttt{None}$
\FOR{$i=N,N-1,\ldots,1$}
    \STATE $t \gets i\Delta t$
    \STATE $\hat{\x}_{0|t}\leftarrow(\x_t-\sigma_t\eps_\Theta(\x_t,t))/\alpha_t$
    \STATE $\D_t\leftarrow\mathcal{C}_t(\hat{\x}_{0|t},\y,\K)$
    \IF{$i<N-N_{\mathrm{warm}}$ and $\D_{\mathrm{prev}}\neq\texttt{None}$}
        \STATE $\widetilde{\D}_t\leftarrow(1-\beta)\D_t+\beta\D_{\mathrm{prev}}$
    \ELSE
        \STATE $\widetilde{\D}_t\leftarrow\D_t$
    \ENDIF
    \STATE $\x_{t-\Delta t}\leftarrow\alpha_{t-\Delta t}\widetilde{\D}_t+\sigma_{t-\Delta t}\eps_\Theta(\x_t,t)$
    \STATE $\D_{\mathrm{prev}}\leftarrow\D_t$
\ENDFOR
\STATE \textbf{return} $\x_{t_0}$
\end{algorithmic}
\end{algorithm}
\FloatBarrier
\subsection{Hyperparameters}
\label{app:hyperparameters}
The following table should be completed with the exact values used for each dataset, task, and backbone.
\small
\begin{longtable}{llcccccccc}
\toprule
\textbf{Dataset} & \textbf{Task} & \textbf{Backbone} & \textbf{NFE} & $\boldsymbol{N_{\mathrm{warm}}}$ & $\boldsymbol{\beta}$ & $\boldsymbol{\mu}$ & $\boldsymbol{\zeta}$ & $\boldsymbol{\eta}$ & $\boldsymbol{\eta_B}$ \\
\midrule
\multicolumn{10}{c}{\textbf{Noiseless} $(\sigma_{\y}=0)$} \\
\midrule
\multirow{8}{*}{FFHQ}
& \multirow{2}{*}{Gaussian deblur} & DiffPIR+LAMP & 100 & 3 & -3.0 & 7.0 & 0.3 & -- & -- \\
& & DDRM+LAMP & 20 & 3 & -0.15 & -- & -- & 0.85 & 1.0 \\
& \multirow{2}{*}{Motion deblur} & DiffPIR+LAMP & 100 & 3 & -3.0 & 7.0 & 0.4 & -- & -- \\
& & DDRM+LAMP & 20 & 3 & -0.15 & -- & -- & 0.85 & 1.0 \\
& \multirow{2}{*}{SR $\times4$} & DiffPIR+LAMP & 100 & 3 & -3.0 & 7.0 & 0.2 & -- & -- \\
& & DDRM+LAMP & 20 & 3 & -0.15 & -- & -- & 0.85 & 1.0 \\
\midrule
\multirow{8}{*}{ImageNet}
& \multirow{2}{*}{Gaussian deblur} & DiffPIR+LAMP & 100 & 3 & -3.0 & 8.0 & 0.3 & -- & -- \\
& & DDRM+LAMP & 20 & 3 & -0.15 & -- & -- & 0.85 & 1.0 \\
& \multirow{2}{*}{Motion deblur} & DiffPIR+LAMP & 100 & 3 & -3.0 & 8.0 & 0.7 & -- & -- \\
& & DDRM+LAMP & 20 & 3 & -0.15 & -- & -- & 0.85 & 1.0 \\
& \multirow{2}{*}{SR $\times4$} & DiffPIR+LAMP & 100 & 3 & -3.0 & 9.0 & 0.5 & -- & -- \\
& & DDRM+LAMP & 20 & 3 & -0.15 & -- & -- & 0.85 & 1.0 \\
\midrule
\multirow{8}{*}{CelebA}
& \multirow{2}{*}{Gaussian deblur} & DiffPIR+LAMP & 100 & 3 & -3.0 & 50.0 & 0.0 & -- & -- \\
& & DDRM+LAMP & 20 & 3 & -0.15 & -- & -- & 0.85 & 1.0 \\
& \multirow{2}{*}{Motion deblur} & DiffPIR+LAMP & 100 & 3 & -3.0 & 8.0 & 0.0 & -- & -- \\
& & DDRM+LAMP & 20 & 3 & -0.15 & -- & -- & 0.85 & 1.0 \\
& \multirow{2}{*}{SR $\times4$} & DiffPIR+LAMP & 100 & 3 & -3.0 & 9.0 & 0.0& -- & -- \\
& & DDRM+LAMP & 20 & 3 & -0.15 & -- & -- & 0.85 & 1.0 \\
\midrule
\multicolumn{10}{c}{\textbf{Noisy} $(\sigma_{\y}=0.05)$} \\
\midrule
\multirow{8}{*}{FFHQ}
& \multirow{2}{*}{Gaussian deblur} & DiffPIR+LAMP & 100 & 3 & -3.0 & 7.0 & 0.3 & -- & -- \\
& & DDRM+LAMP & 20 & 3 & -0.15 & -- & -- & 0.85 & 1.0 \\
& \multirow{2}{*}{Motion deblur} & DiffPIR+LAMP & 100 & 3 & -3.0 & 7.0 & 0.4 & -- & -- \\
& & DDRM+LAMP & 20 & 3 & -0.15 & -- & -- & 0.85 & 1.0 \\
& \multirow{2}{*}{SR $\times4$} & DiffPIR+LAMP & 100 & 3 & -3.0 & 7.0 & 0.2 & -- & -- \\
& & DDRM+LAMP & 20 & 3 & -0.15 & -- & -- & 0.85 & 1.0 \\
\midrule
\multirow{8}{*}{ImageNet}
& \multirow{2}{*}{Gaussian deblur} & DiffPIR+LAMP & 100 & 3 & -3.0 & 8.0 & 0.3 & -- & -- \\
& & DDRM+LAMP & 20 & 3 & -0.15 & -- & -- & 0.85 & 1.0 \\
& \multirow{2}{*}{Motion deblur} & DiffPIR+LAMP & 100 & 3 & -3.0 & 8.0 & 0.7 & -- & -- \\
& & DDRM+LAMP & 20 & 3 & -0.15 & -- & -- & 0.85 & 1.0 \\
& \multirow{2}{*}{SR $\times4$} & DiffPIR+LAMP & 100 & 3 & -3.0 & 9.0 & 0.5 & -- & -- \\
& & DDRM+LAMP & 20 & 3 & -0.15 & -- & -- & 0.85 & 1.0 \\
\midrule
\multirow{8}{*}{CelebA}
& \multirow{2}{*}{Gaussian deblur} & DiffPIR+LAMP & 100 & 3 & -3.0 & 50.0 & 0.0 & -- & -- \\
& & DDRM+LAMP & 20 & 3 & -0.15 & -- & -- & 0.85 & 1.0 \\
& \multirow{2}{*}{Motion deblur} & DiffPIR+LAMP & 100 & 3 & -3.0 & 8.0 & 0.0 & -- & -- \\
& & DDRM+LAMP & 20 & 3 & -0.15 & -- & -- & 0.85 & 1.0 \\
& \multirow{2}{*}{SR $\times4$} & DiffPIR+LAMP & 100 & 3 & -3.0 & 9.0 & 0.0& -- & -- \\
& & DDRM+LAMP & 20 & 3 & -0.15 & -- & -- & 0.85 & 1.0 \\
\bottomrule
\caption{Hyperparameters used by LAMP in the noiseless setting $(\sigma_{\y}=0)$ and noisy setting $(\sigma_{\y}=0.05)$. The parameters $\mu$ and $\zeta$ are used by DiffPIR, while $\eta$ and $\eta_B$ are used by DDRM.}
\label{tab:app_hyperparameters}
\end{longtable}
\FloatBarrier
\subsection{Neural network architecture}
\label{app:network_architecture}
All experiments use a pretrained DDPM denoiser with the \texttt{score\_sde} UNet architecture. The checkpoint is automatically detected by the model loader and loaded through the \texttt{DDPMModel} implementation in \texttt{sde\_model.py}, rather than through the \texttt{guided\_diffusion} UNet implementation. The model predicts the diffusion noise $\eps_\Theta(\x_t,t)$ and is not class-conditional.
\begin{table}[h]
\centering
\small
\begin{tabular}{lc}
\toprule
\textbf{Parameter} & \textbf{Value} \\
\midrule
Image size & $256 \times 256$ \\
Base channels & $128$ \\
Residual blocks per level & $2$ \\
Head channels & $64$ \\
Attention resolutions & $16$ \\
Channel multipliers & $(1,1,2,2,4,4)$ \\
Dropout & $0.0$ \\
\bottomrule
\end{tabular}
\caption{UNet architecture used by the pretrained \texttt{score\_sde} DDPM denoiser.}
\label{tab:app_network_architecture}
\end{table}
\FloatBarrier
\subsection{Diffusion model and sampling schedule}
\label{app:diffusion_schedule}
The denoiser is a Ho et al.-style DDPM model trained to predict only the noise component, rather than both noise and variance. Accordingly, we use
\begin{equation}
    \mathrm{learn\_sigma}=\texttt{false},
\end{equation}
so that the sampling code uses the fixed small variance parameterization instead of the learned-range variance parameterization. The model uses a linear diffusion schedule with $1000$ training diffusion steps. At inference time, we use a respaced deterministic reverse schedule with $100$ selected timesteps.
\begin{table}[h]
\centering
\small
\begin{tabular}{lc}
\toprule
\textbf{Parameter} & \textbf{Value} \\
\midrule
Learned variance & No \\
Number of training diffusion steps & $1000$ \\
Noise schedule & Linear \\
\bottomrule
\end{tabular}
\caption{Diffusion-model configuration used by the pretrained denoiser.}
\label{tab:app_diffusion_schedule}
\end{table}
\subsection{Datasets and preprocessing}
\label{app:datasets}
We evaluate on FFHQ, ImageNet, and CelebA, as in the main paper. The following details should be completed before submission.
\begin{table}[h]
\centering
\scriptsize
\begin{tabular}{lcccc}
\toprule
\textbf{Dataset} & \textbf{Resolution} & \textbf{Number of test images} & \textbf{Crop/resize} & \textbf{Pretrained model} \\
\midrule
FFHQ & $256 \times 256$ & $100$ & \texttt{None} & \texttt{diffusion\_ffhq\_10m.pt} \citep{dhariwal2021diffusion} \\
ImageNet & $256 \times 256$ & $100$ & \texttt{None} & \texttt{256x256\_diffusion\_uncond.pt} \citep{dhariwal2021diffusion} \\
CelebA & $256 \times 256$ & $100$ & \texttt{None} & \texttt{celeba\_hq.ckpt} \citep{dhariwal2021diffusion} \\
\bottomrule
\end{tabular}
\caption{Dataset and preprocessing details.}
\label{tab:app_datasets}
\end{table}
\FloatBarrier
\subsection{Forward operators}
\label{app:forward_operators}
We consider the following linear inverse problems.
\begin{table}[h]
\centering
\small
\begin{tabular}{lcc}
\toprule
\textbf{Task} & \textbf{Operator} & \textbf{Parameters} \\
\midrule
Gaussian deblurring & Gaussian convolution & \texttt{kernel\_size} $61\times61$, \texttt{sigma} $3.0$ \\
Motion deblurring & Motion convolution & \texttt{kernel\_size} $61\times61$, \texttt{intensity} $0.5$ \\
Super-resolution & Average Downsampling & \texttt{downsampling\_type} average, \texttt{factor} 4 \\
\bottomrule
\end{tabular}
\caption{Forward operators used in the experiments.}
\label{tab:app_forward_operators}
\end{table}
\FloatBarrier
\section{Additional Quantitative Results}
\label{app:additional_experiments}
This section presents additional quantitative results that complement the paper.
While the paper presents the main comparisons of PSNR, SSIM, and LPIPS mean values for noisy settings, here we consider experiments in noiseless settings, as well as ablations on additional degradation levels, confidence intervals, and per-task breakdowns.
\subsection{Noiseless Case}\label{app:noiseless_experiments}
{\centering
\small
\begin{longtable}{lccccccc}
\toprule
\multicolumn{1}{c}{\textbf{FFHQ}} & \multicolumn{1}{c}{\textbf{}} & \multicolumn{2}{c}{\textbf{Deblur (Gaussian)}} & \multicolumn{2}{c}{\textbf{Deblur (Motion)}} & \multicolumn{2}{c}{\textbf{SR ($\times 4$)}} \\
\cmidrule(lr){3-4} \cmidrule(lr){5-6} \cmidrule(lr){7-8}
\textbf{Method} & NFE & PSNR$\uparrow$ & SSIM$\uparrow$ & PSNR$\uparrow$ & SSIM$\uparrow$ & PSNR$\uparrow$ & SSIM$\uparrow$ \\
\midrule
DPS \citep{chung2022dps} & 1000 & \textbf{26.32} & 0.741 & 22.69 & 0.636 & 24.68 & 0.689 \\
DiffPIR \citep{zhu2023diffpir} & 100 & 25.10 & 0.779 & 24.51 & 0.767 & 28.60 & 0.837 \\
DDRM \citep{kawar2022ddrm} & 20 & 25.53 & \underline{0.796} & \underline{24.69} & 0.773 & \underline{29.08} & 0.843 \\
DiffPIR-1M & 100 & 24.51 & 0.699 & 24.17 & 0.744 & 26.50 & 0.720 \\
DiffPIR-2M & 100 & 24.26 & 0.684 & 23.95 & 0.722 & 26.09 & 0.707 \\
DDRM-1M & 20 & 24.87 & 0.688 & 24.40 & \underline{0.777} & 26.98 & 0.738 \\
DDRM-2M & 20 & 24.85 & 0.691 & 24.33 & 0.772 & 26.82 & 0.732 \\
\noalign{\vskip 2pt}
\cdashline{1-8}
\noalign{\vskip 2pt}
DiffPIR+LAMP (Ours) & 100 & 25.28 & 0.790 & 24.62 & 0.775 & 29.01 & \textbf{0.850} \\
DDRM+LAMP (Ours) & 20 & \underline{25.59} & \textbf{0.802} & \textbf{24.75} & \textbf{0.784} & \textbf{29.23} & \underline{0.849} \\
\midrule
\multicolumn{1}{c}{\textbf{ImageNet}} & \multicolumn{1}{c}{\textbf{}} & \multicolumn{2}{c}{\textbf{Deblur (Gaussian)}} & \multicolumn{2}{c}{\textbf{Deblur (Motion)}} & \multicolumn{2}{c}{\textbf{SR ($\times 4$)}} \\
\cmidrule(lr){3-4} \cmidrule(lr){5-6} \cmidrule(lr){7-8}
\textbf{Method} & NFE & PSNR$\uparrow$ & SSIM$\uparrow$ & PSNR$\uparrow$ & SSIM$\uparrow$ & PSNR$\uparrow$ & SSIM$\uparrow$ \\
\midrule
DPS \citep{chung2022dps} & 1000 & 21.66 & 0.534 & 17.29 & 0.388 & 20.94 & 0.484 \\
DiffPIR \citep{zhu2023diffpir} & 100 & 22.18 & 0.639 & 21.57 & 0.614 & \underline{24.64} & \underline{0.719} \\
DDRM \citep{kawar2022ddrm} & 20 & \underline{22.58} & \underline{0.651} & \underline{21.82} & \underline{0.619} & 24.35 & 0.692 \\
DiffPIR-1M & 100 & 21.34 & 0.579 & 20.78 & 0.562 & 21.77 & 0.562 \\
DiffPIR-2M & 100 & 20.81 & 0.541 & 20.50 & 0.534 & 20.98 & 0.517 \\
DDRM-1M & 20 & 22.29 & 0.651 & 21.09 & 0.610 & 22.42 & 0.599 \\
DDRM-2M & 20 & 22.23 & 0.646 & 21.02 & 0.604 & 22.24 & 0.587 \\
\noalign{\vskip 2pt}
\cdashline{1-8}
\noalign{\vskip 2pt}
DiffPIR+LAMP (Ours) & 100 & 22.34 & 0.648 & 21.62 & 0.617 & \textbf{24.80} & \textbf{0.726} \\
DDRM+LAMP (Ours) & 20 & \textbf{22.61} & \textbf{0.655} & \textbf{21.85} & \textbf{0.627} & 24.43 & 0.697 \\
\midrule
\multicolumn{1}{c}{\textbf{CelebA}} & \multicolumn{1}{c}{\textbf{}} & \multicolumn{2}{c}{\textbf{Deblur (Gaussian)}} & \multicolumn{2}{c}{\textbf{Deblur (Motion)}} & \multicolumn{2}{c}{\textbf{SR ($\times 4$)}} \\
\cmidrule(lr){3-4} \cmidrule(lr){5-6} \cmidrule(lr){7-8}
\textbf{Method} & NFE & PSNR$\uparrow$ & SSIM$\uparrow$ & PSNR$\uparrow$ & SSIM$\uparrow$ & PSNR$\uparrow$ & SSIM$\uparrow$ \\
\midrule
DPS \citep{chung2022dps} & 1000 & \textbf{27.36} & 0.750 & 25.02 & 0.697 & 26.43 & 0.725 \\
DiffPIR \citep{zhu2023diffpir} & 100 & 25.30 & 0.745 & 24.86 & 0.754 & 28.56 & 0.818 \\
DDRM \citep{kawar2022ddrm} & 20 & 26.06 & \underline{0.804} & \underline{25.49} & \underline{0.795} & \underline{29.96} & \underline{0.847} \\
DiffPIR-1M & 100 & 25.30 & 0.745 & 24.86 & 0.754 & 28.56 & 0.818 \\
DiffPIR-2M & 100 & 25.07 & 0.729 & 24.71 & 0.740 & 28.20 & 0.805 \\
DDRM-1M & 20 & 25.86 & 0.793 & 25.05 & 0.783 & 28.80 & 0.825 \\
DDRM-2M & 20 & 25.79 & 0.788 & 24.99 & 0.778 & 28.58 & 0.817 \\
\noalign{\vskip 2pt}
\cdashline{1-8}
\noalign{\vskip 2pt}
DiffPIR+LAMP (Ours) & 100 & 25.47 & 0.756 & 24.97 & 0.763 & 28.82 & 0.827 \\
DDRM+LAMP (Ours) & 20 & \underline{26.11} & \textbf{0.808} & \textbf{25.53} & \textbf{0.800} & \textbf{30.15} & \textbf{0.854} \\
\bottomrule
\caption{Quantitative comparison in the noiseless setting. Best values are in bold and second-best values are underlined.}
\label{tab:main_results_noiseless}
\end{longtable}}
\FloatBarrier
\subsection{Robustness to Increasing Measurement Noise}\label{app:noise_sweep}
In this section, we evaluate the robustness of the considered methods under increasing measurement noise. We report results on FFHQ for Gaussian deblurring and motion deblurring, varying the noise standard deviation $\sigma \in \{0.01, 0.03, 0.05, 0.07, 0.10\}$. This analysis complements the main results by showing how reconstruction quality degrades as the inverse problem becomes progressively more corrupted by noise.
\begin{table*}[bthp]
\centering
\setlength{\tabcolsep}{3.5pt}
\scriptsize
\begin{tabular}{lcccccccccc}
\toprule
\multicolumn{1}{c}{\textbf{FFHQ}}
& \multicolumn{10}{c}{\textbf{Deblur (Gaussian)}} \\
\cmidrule(lr){2-11}
\textbf{Method}
& \multicolumn{2}{c}{$\sigma=0.01$}
& \multicolumn{2}{c}{$\sigma=0.03$}
& \multicolumn{2}{c}{$\sigma=0.05$}
& \multicolumn{2}{c}{$\sigma=0.07$}
& \multicolumn{2}{c}{$\sigma=0.10$} \\
\cmidrule(lr){2-3} \cmidrule(lr){4-5} \cmidrule(lr){6-7} \cmidrule(lr){8-9} \cmidrule(lr){10-11}
& PSNR$\uparrow$ & SSIM$\uparrow$
& PSNR$\uparrow$ & SSIM$\uparrow$
& PSNR$\uparrow$ & SSIM$\uparrow$
& PSNR$\uparrow$ & SSIM$\uparrow$
& PSNR$\uparrow$ & SSIM$\uparrow$ \\
\midrule
DiffPIR
& 25.08 & 0.774 & 24.94 & 0.751 & 24.79 & 0.737 & 24.64 & 0.725 & 24.44 & 0.712 \\
DDRM
& \underline{25.52} & \underline{0.799} & \underline{25.54} & \underline{0.785} & \underline{25.51} & \underline{0.776} & \underline{25.45} & \underline{0.768} & \underline{25.33} & \underline{0.758} \\
DiffPIR-1M
& 24.36 & 0.680 & 24.04 & 0.644 & 23.79 & 0.623 & 23.59 & 0.607 & 23.32 & 0.588 \\
DiffPIR-2M
& 24.05 & 0.658 & 23.68 & 0.620 & 23.42 & 0.598 & 23.20 & 0.582 & 22.93 & 0.564 \\
DDRM-1M
& 12.71 & 0.092 & 12.09 & 0.079 & 11.80 & 0.073 & 11.60 & 0.069 & 11.33 & 0.063 \\
DDRM-2M
& 12.78 & 0.093 & 12.14 & 0.080 & 11.85 & 0.074 & 11.65 & 0.069 & 11.38 & 0.064 \\
\noalign{\vskip 2pt}
\cdashline{1-11}
\noalign{\vskip 2pt}
DiffPIR-LAMP (Ours)
& 25.26 & 0.786 & 25.15 & 0.764 & 25.01 & 0.750 & 24.87 & 0.739 & 24.67 & 0.726 \\
DDRM-LAMP (Ours)
& \textbf{25.57} & \textbf{0.802} & \textbf{25.58} & \textbf{0.789} & \textbf{25.55} & \textbf{0.779} & \textbf{25.49} & \textbf{0.771} & \textbf{25.38} & \textbf{0.761} \\
\midrule
\multicolumn{1}{c}{\textbf{FFHQ}}
& \multicolumn{10}{c}{\textbf{Deblur (Motion)}} \\
\cmidrule(lr){2-11}
\textbf{Method}
& \multicolumn{2}{c}{$\sigma=0.01$}
& \multicolumn{2}{c}{$\sigma=0.03$}
& \multicolumn{2}{c}{$\sigma=0.05$}
& \multicolumn{2}{c}{$\sigma=0.07$}
& \multicolumn{2}{c}{$\sigma=0.10$} \\
\cmidrule(lr){2-3} \cmidrule(lr){4-5} \cmidrule(lr){6-7} \cmidrule(lr){8-9} \cmidrule(lr){10-11}
& PSNR$\uparrow$ & SSIM$\uparrow$
& PSNR$\uparrow$ & SSIM$\uparrow$
& PSNR$\uparrow$ & SSIM$\uparrow$
& PSNR$\uparrow$ & SSIM$\uparrow$
& PSNR$\uparrow$ & SSIM$\uparrow$ \\
\midrule
DiffPIR
& 24.47 & 0.758 & 24.00 & 0.714 & 23.50 & 0.688 & 23.09 & 0.670 & 22.59 & 0.650 \\
DDRM
& \underline{24.69} & \underline{0.779} & \underline{24.69} & \underline{0.750} & \underline{24.48} & \underline{0.732} & \underline{24.25} & \underline{0.719} & \underline{23.93} & \underline{0.704} \\
DiffPIR-1M
& 23.10 & 0.549 & 22.08 & 0.452 & 21.54 & 0.426 & 21.14 & 0.414 & 20.69 & 0.403 \\
DiffPIR-2M
& 22.88 & 0.534 & 21.83 & 0.438 & 21.27 & 0.412 & 20.87 & 0.399 & 20.41 & 0.387 \\
DDRM-1M
& 19.08 & 0.256 & 13.38 & 0.094 & 11.51 & 0.061 & 10.50 & 0.046 & 9.70 & 0.036 \\
DDRM-2M
& 19.21 & 0.262 & 13.47 & 0.095 & 11.58 & 0.062 & 10.57 & 0.047 & 9.76 & 0.036 \\
\noalign{\vskip 2pt}
\cdashline{1-11}
\noalign{\vskip 2pt}
DiffPIR-LAMP (Ours)
& 24.59 & 0.768 & 24.16 & 0.725 & 23.68 & 0.700 & 23.27 & 0.682 & 22.77 & 0.663 \\
DDRM-LAMP (Ours)
& \textbf{24.72} & \textbf{0.785} & \textbf{24.73} & \textbf{0.756} & \textbf{24.52} & \textbf{0.737} & \textbf{24.29} & \textbf{0.724} & \textbf{23.97} & \textbf{0.708} \\
\bottomrule
\end{tabular}
\caption{Robustness to increasing measurement noise on FFHQ. We report PSNR and SSIM for Gaussian and motion deblurring under different noise levels $\sigma$. Best values are in bold and second-best values are underlined.}
\label{tab:appendix_noise_sweep}
\end{table*}
\FloatBarrier
\subsection{Robustness Across Multiple Runs}\label{app:robustness_std}
To assess the robustness of the evaluated methods, we repeat each experiment across multiple independent runs and analyze the variability of the results. In particular, for each run we compute the average PSNR and SSIM over the dataset, and we then report the standard deviation of these averages across runs. This provides a measure of stability with respect to stochasticity in the sampling process and implementation details. The results in Table~\ref{tab:appendix_robustness_noisy} complement the main quantitative comparison by highlighting not only performance but also consistency across repeated executions.
\begin{table*}[h!]
\centering
\setlength{\tabcolsep}{3.5pt}
\scriptsize
\begin{tabular}{lccccccc}
\toprule
\multicolumn{1}{c}{\textbf{CelebA}} & \multicolumn{1}{c}{} & \multicolumn{2}{c}{\textbf{Deblur (Gaussian)}} & \multicolumn{2}{c}{\textbf{Deblur (Motion)}} & \multicolumn{2}{c}{\textbf{SR ($\times 4$)}} \\
\cmidrule(lr){3-4} \cmidrule(lr){5-6} \cmidrule(lr){7-8}
\textbf{Method} & NFE & PSNR$\uparrow$ & SSIM$\uparrow$ & PSNR$\uparrow$ & SSIM$\uparrow$ & PSNR$\uparrow$ & SSIM$\uparrow$ \\
\midrule
DPS \citep{chung2022dps} & 1000 & \textbf{26.45}\std{0.036} & 0.720\std{0.0012} & 23.74\std{0.055} & 0.659\std{0.0007} & 25.88\std{0.018} & 0.710\std{0.0009} \\
DiffPIR \citep{zhu2023diffpir} & 100 & 24.84\std{0.044} & 0.705\std{0.0013} & 23.20\std{0.045} & 0.567\std{0.0025} & 26.90\std{0.024} & 0.690\std{0.0009} \\
DDRM \citep{kawar2022ddrm} & 20 & 26.11\std{0.018} & \underline{0.789}\std{0.0006} & \underline{25.52}\std{0.021} & \underline{0.770}\std{0.0005} & \underline{28.97}\std{0.025} & \underline{0.818}\std{0.0005} \\
DiffPIR-1M & 100 & 24.84\std{0.044} & 0.705\std{0.0013} & 23.20\std{0.045} & 0.567\std{0.0025} & 26.90\std{0.024} & 0.690\std{0.0009} \\
DiffPIR-2M & 100 & 24.56\std{0.046} & 0.687\std{0.0012} & 23.06\std{0.046} & 0.559\std{0.0027} & 26.59\std{0.024} & 0.675\std{0.0009} \\
DDRM-1M & 20 & 15.51\std{0.017} & 0.146\std{0.0003} & 14.37\std{0.009} & 0.108\std{0.0003} & 27.92\std{0.022} & 0.794\std{0.0008} \\
DDRM-2M & 20 & 15.61\std{0.017} & 0.149\std{0.0003} & 14.51\std{0.009} & 0.111\std{0.0003} & 27.76\std{0.022} & 0.789\std{0.0009} \\
\noalign{\vskip 2pt}
\cdashline{1-8}
\noalign{\vskip 2pt}
DiffPIR-LAMP (Ours) & 100 & 25.04\std{0.043} & 0.716\std{0.0013} & 23.26\std{0.042} & 0.566\std{0.0022} & 27.11\std{0.024} & 0.699\std{0.0009} \\
DDRM-LAMP (Ours) & 20 & \underline{26.14}\std{0.018} & \textbf{0.791}\std{0.0006} & \textbf{25.55}\std{0.021} & \textbf{0.772}\std{0.0005} & \textbf{29.12}\std{0.026} & \textbf{0.823}\std{0.0005} \\
\midrule
\multicolumn{1}{c}{\textbf{FFHQ}} & \multicolumn{1}{c}{} & \multicolumn{2}{c}{\textbf{Deblur (Gaussian)}} & \multicolumn{2}{c}{\textbf{Deblur (Motion)}} & \multicolumn{2}{c}{\textbf{SR ($\times 4$)}} \\
\cmidrule(lr){3-4} \cmidrule(lr){5-6} \cmidrule(lr){7-8}
\textbf{Method} & NFE & PSNR$\uparrow$ & SSIM$\uparrow$ & PSNR$\uparrow$ & SSIM$\uparrow$ & PSNR$\uparrow$ & SSIM$\uparrow$ \\
\midrule
DPS \citep{chung2022dps} & 1000 & 25.35\std{0.046} & 0.706\std{0.0019} & 21.58\std{0.077} & 0.595\std{0.0035} & 24.27\std{0.032} & 0.676\std{0.0010} \\
DiffPIR \citep{zhu2023diffpir} & 100 & 24.84\std{0.026} & 0.741\std{0.0009} & 23.66\std{0.019} & 0.696\std{0.0010} & 26.54\std{0.014} & 0.676\std{0.0005} \\
DDRM \citep{kawar2022ddrm} & 20 & \underline{25.51}\std{0.024} & \underline{0.776}\std{0.0006} & \underline{24.48}\std{0.010} & \underline{0.732}\std{0.0005} & \underline{27.94}\std{0.029} & \underline{0.804}\std{0.0005} \\
DiffPIR-1M & 100 & 23.71\std{0.049} & 0.610\std{0.0056} & 21.23\std{0.028} & 0.386\std{0.0015} & 25.01\std{0.094} & 0.592\std{0.0047} \\
DiffPIR-2M & 100 & 23.35\std{0.058} & 0.585\std{0.0058} & 21.00\std{0.035} & 0.376\std{0.0018} & 24.65\std{0.091} & 0.574\std{0.0044} \\
DDRM-1M & 20 & 11.80\std{0.011} & 0.073\std{0.0001} & 11.51\std{0.005} & 0.061\std{0.0002} & 26.11\std{0.103} & 0.696\std{0.0064} \\
DDRM-2M & 20 & 11.85\std{0.012} & 0.074\std{0.0001} & 11.58\std{0.005} & 0.062\std{0.0002} & 25.99\std{0.100} & 0.693\std{0.0064} \\
\noalign{\vskip 2pt}
\cdashline{1-8}
\noalign{\vskip 2pt}
DiffPIR-LAMP (Ours) & 100 & 25.18\std{0.023} & 0.761\std{0.0007} & 23.94\std{0.016} & 0.713\std{0.0008} & 27.05\std{0.010} & 0.704\std{0.0004} \\
DDRM-LAMP (Ours) & 20 & \textbf{25.55}\std{0.023} & \textbf{0.779}\std{0.0005} & \textbf{24.52}\std{0.009} & \textbf{0.737}\std{0.0004} & \textbf{28.07}\std{0.028} & \textbf{0.808}\std{0.0005} \\
\midrule
\multicolumn{1}{c}{\textbf{ImageNet}} & \multicolumn{1}{c}{} & \multicolumn{2}{c}{\textbf{Deblur (Gaussian)}} & \multicolumn{2}{c}{\textbf{Deblur (Motion)}} & \multicolumn{2}{c}{\textbf{SR ($\times 4$)}} \\
\cmidrule(lr){3-4} \cmidrule(lr){5-6} \cmidrule(lr){7-8}
\textbf{Method} & NFE & PSNR$\uparrow$ & SSIM$\uparrow$ & PSNR$\uparrow$ & SSIM$\uparrow$ & PSNR$\uparrow$ & SSIM$\uparrow$ \\
\midrule
DPS \citep{chung2022dps} & 1000 & 21.02\std{0.198} & 0.495\std{0.0063} & 16.97\std{0.110} & 0.367\std{0.0069} & 20.76\std{0.074} & 0.477\std{0.0047} \\
DiffPIR \citep{zhu2023diffpir} & 100 & 21.77\std{0.044} & 0.581\std{0.0019} & 20.96\std{0.024} & 0.528\std{0.0015} & 23.17\std{0.027} & 0.569\std{0.0007} \\
DDRM \citep{kawar2022ddrm} & 20 & \underline{22.38}\std{0.006} & \underline{0.611}\std{0.0004} & \underline{21.36}\std{0.013} & \underline{0.544}\std{0.0002} & \underline{23.73}\std{0.042} & \underline{0.642}\std{0.0014} \\
DiffPIR-1M & 100 & 19.27\std{0.089} & 0.361\std{0.0047} & 13.92\std{0.018} & 0.106\std{0.0014} & 20.85\std{0.109} & 0.439\std{0.0034} \\
DiffPIR-2M & 100 & 18.73\std{0.099} & 0.340\std{0.0045} & 13.46\std{0.022} & 0.100\std{0.0016} & 20.02\std{0.102} & 0.392\std{0.0024} \\
DDRM-1M & 20 & 8.53\std{0.006} & 0.043\std{0.0002} & 8.71\std{0.005} & 0.039\std{0.0002} & 21.86\std{0.047} & 0.548\std{0.0038} \\
DDRM-2M & 20 & 8.50\std{0.006} & 0.043\std{0.0002} & 8.68\std{0.005} & 0.039\std{0.0002} & 21.70\std{0.050} & 0.539\std{0.0038} \\
\noalign{\vskip 2pt}
\cdashline{1-8}
\noalign{\vskip 2pt}
DiffPIR-LAMP (Ours) & 100 & 22.01\std{0.030} & 0.596\std{0.0010} & 21.01\std{0.023} & 0.529\std{0.0014} & 23.43\std{0.024} & 0.588\std{0.0005} \\
DDRM-LAMP (Ours) & 20 & \textbf{22.41}\std{0.005} & \textbf{0.613}\std{0.0004} & \textbf{21.38}\std{0.012} & \textbf{0.547}\std{0.0002} & \textbf{23.80}\std{0.042} & \textbf{0.647}\std{0.0014} \\
\bottomrule
\end{tabular}
\caption{Robustness evaluation in the noisy setting ($\sigma_{\y} = 0.05$) across multiple runs. For each method and task, we report the average PSNR and SSIM, together with the standard deviation computed over $R=5$ runs with different seeds, indicated in gray. Best values are in bold and second-best values are underlined.}
\label{tab:appendix_robustness_noisy}
\end{table*}
\FloatBarrier
\section{Additional Qualitative Results}
\label{app:qualitative}
This section contains additional visual comparisons (selected by best LPIPS).
\begin{figure}[h!]
  \setlength{\tabcolsep}{2pt}
  \centering
  \scriptsize
\newcommand{\imgw}{0.12\linewidth}
\newcommand{\imgrow}[7]{
  \includegraphics[width=\imgw]{#1} &
  \includegraphics[width=\imgw]{#2} &
  \includegraphics[width=\imgw]{#3} &
  \includegraphics[width=\imgw]{#4} &
  \includegraphics[width=\imgw]{#5} &
  \includegraphics[width=\imgw]{#6} &
  \includegraphics[width=\imgw]{#7} \\
}
  \resizebox{0.85\linewidth}{!}{
  \begin{tabular}{@{}ccccccc@{}}
    Measured & DPS & DiffPIR & DDRM & DiffPIR+LAMP & DDRM+LAMP & Original \\
    \imgrow{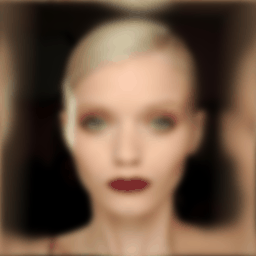}{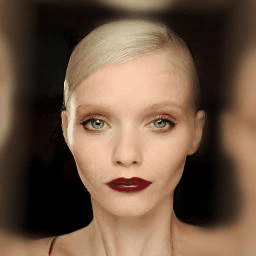}{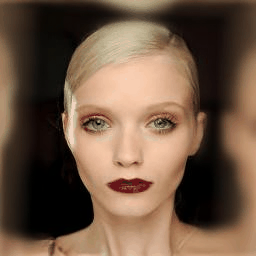}{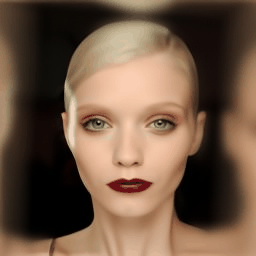}{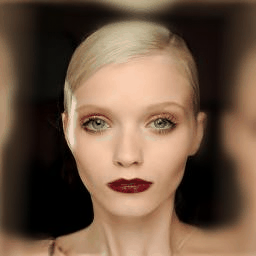}{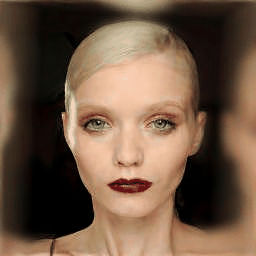}{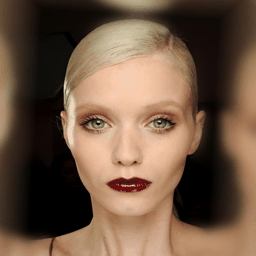}
    \imgrow{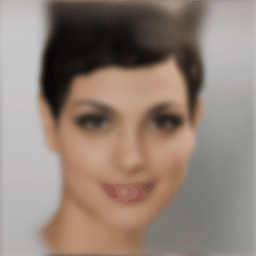}{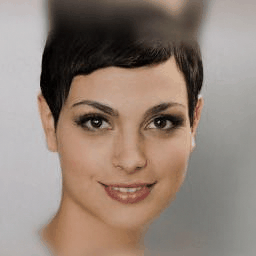}{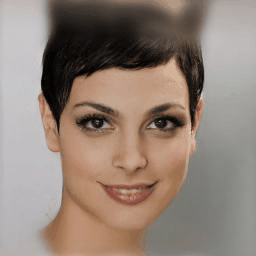}{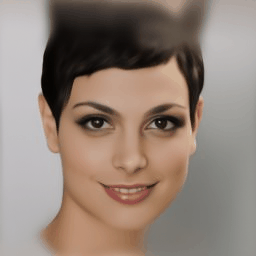}{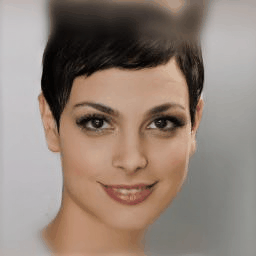}{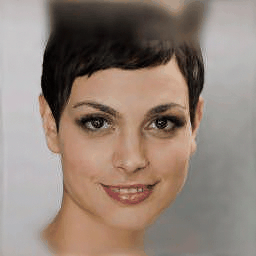}{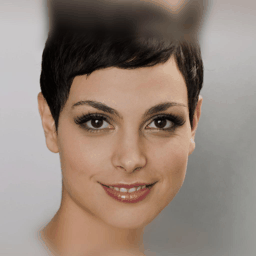}
    \imgrow{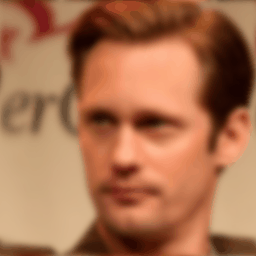}{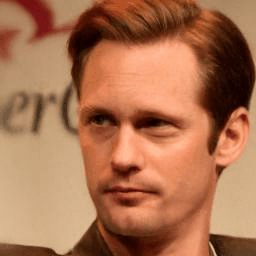}{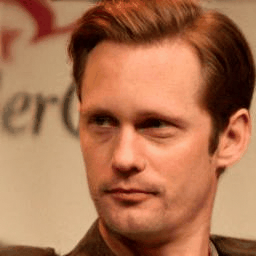}{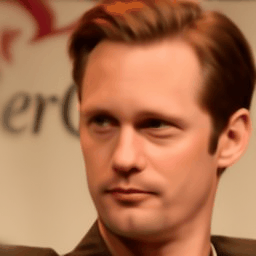}{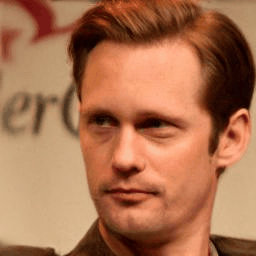}{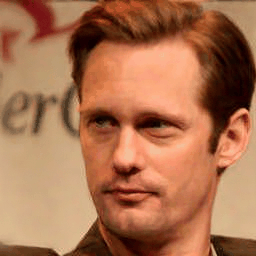}{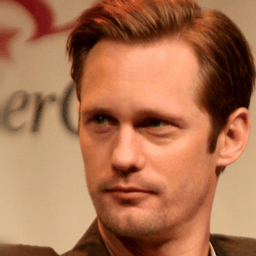}
    \imgrow{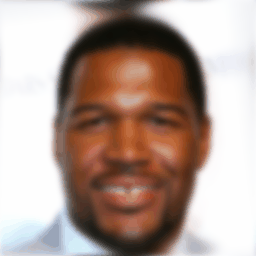}{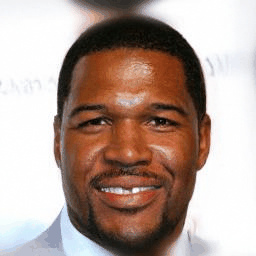}{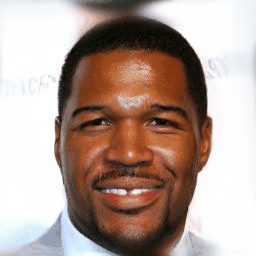}{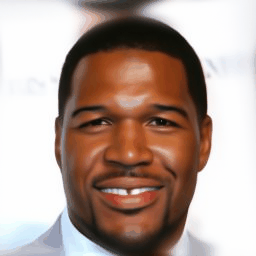}{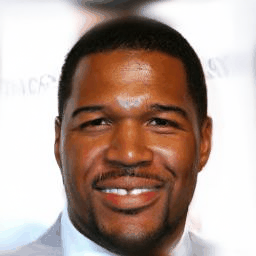}{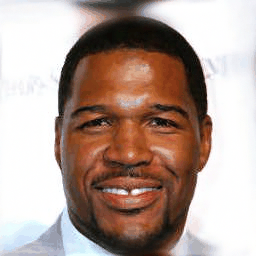}{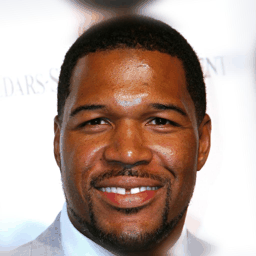}
    \imgrow{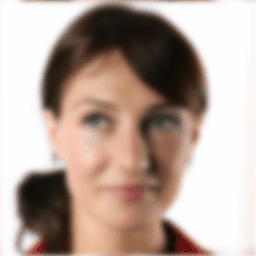}{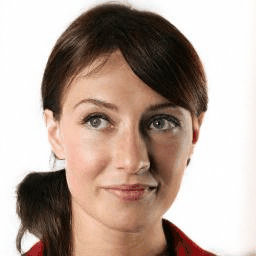}{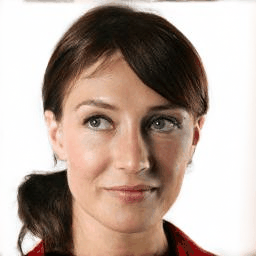}{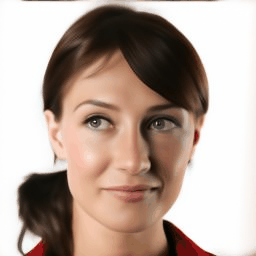}{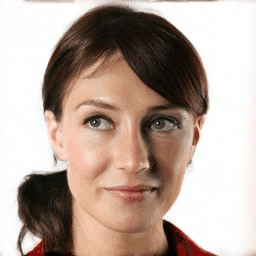}{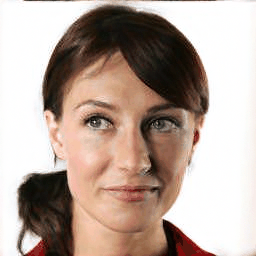}{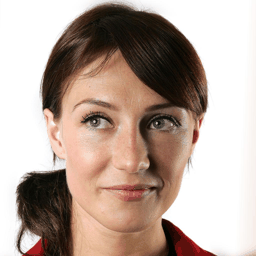}
  \end{tabular}}
  \caption{Qualitative comparison on CelebA -- Gaussian deblurring (noiseless).}
  \label{fig:celeba_gauss_noiseless_lpips}
\end{figure}
\FloatBarrier
\begin{figure}[h!]
  \setlength{\tabcolsep}{2pt}
  \centering
  \scriptsize
\newcommand{\imgw}{0.12\linewidth}
\newcommand{\imgrow}[7]{
  \includegraphics[width=\imgw]{#1} &
  \includegraphics[width=\imgw]{#2} &
  \includegraphics[width=\imgw]{#3} &
  \includegraphics[width=\imgw]{#4} &
  \includegraphics[width=\imgw]{#5} &
  \includegraphics[width=\imgw]{#6} &
  \includegraphics[width=\imgw]{#7} \\
}
  \resizebox{0.85\linewidth}{!}{
  \begin{tabular}{@{}ccccccc@{}}
    Measured & DPS & DiffPIR & DDRM & DiffPIR+LAMP & DDRM+LAMP & Original \\
    \imgrow{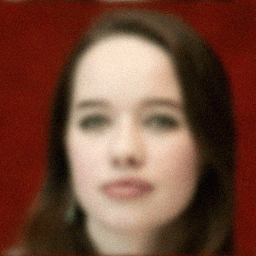}{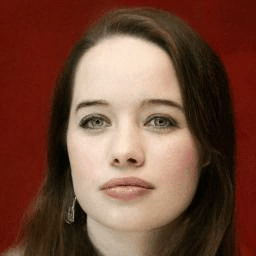}{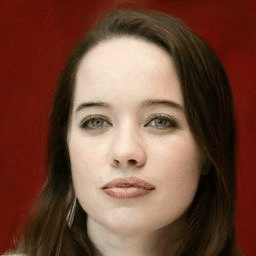}{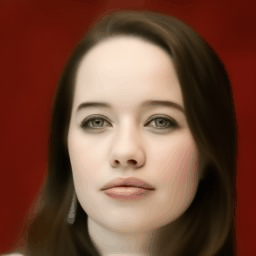}{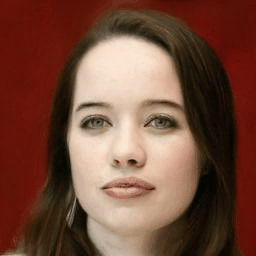}{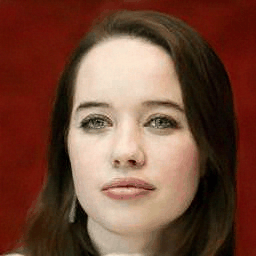}{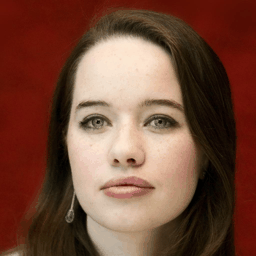}
    \imgrow{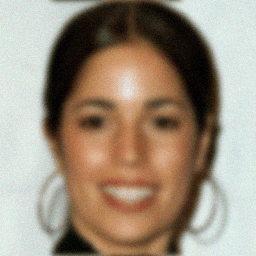}{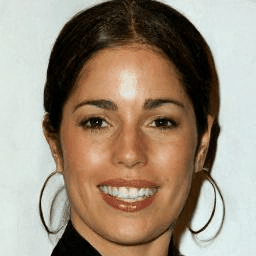}{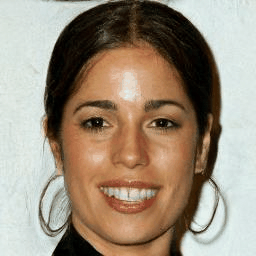}{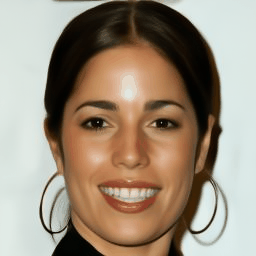}{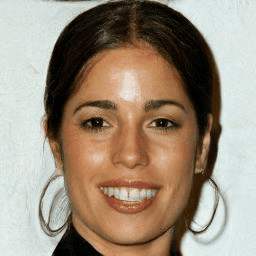}{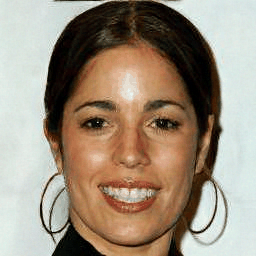}{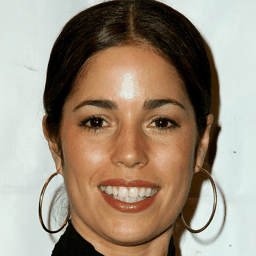}
    \imgrow{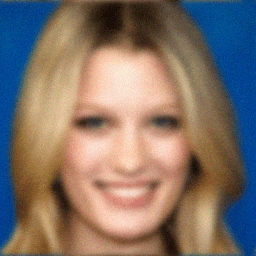}{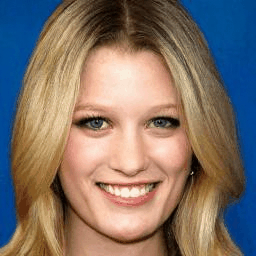}{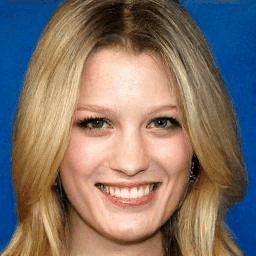}{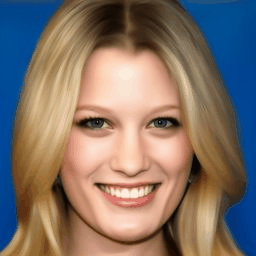}{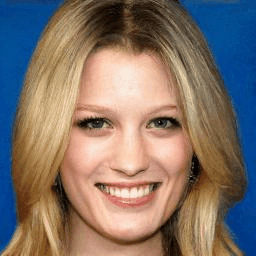}{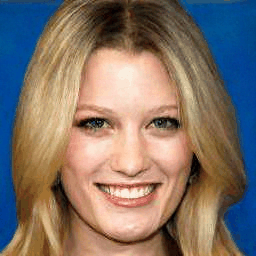}{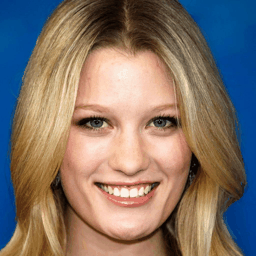}
    \imgrow{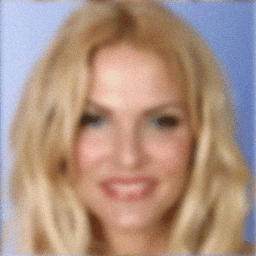}{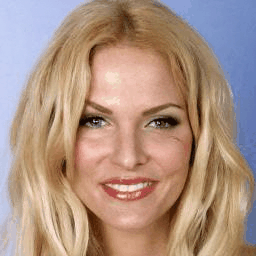}{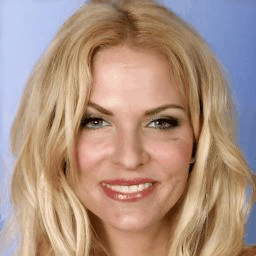}{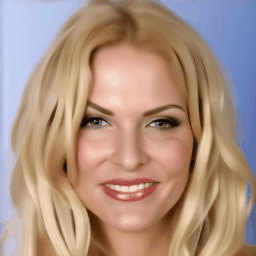}{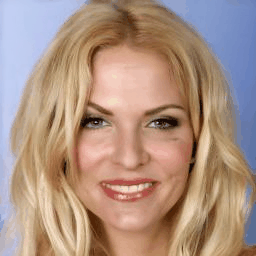}{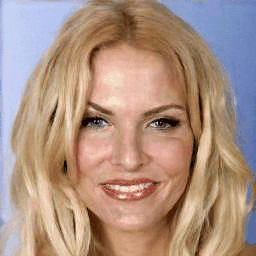}{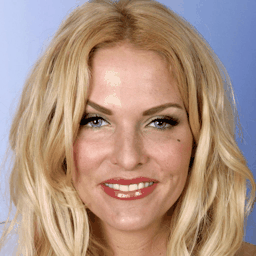}
    \imgrow{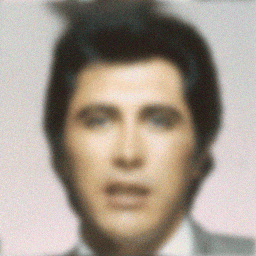}{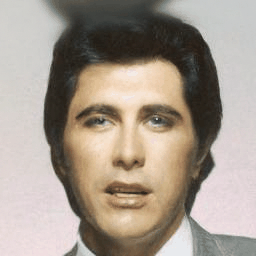}{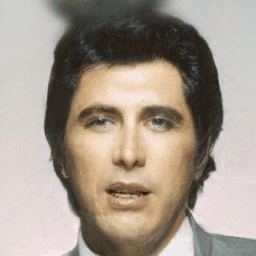}{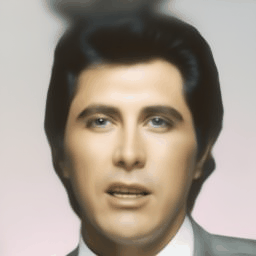}{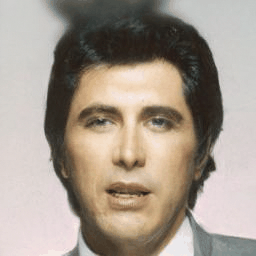}{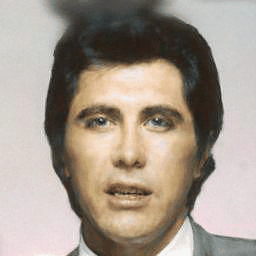}{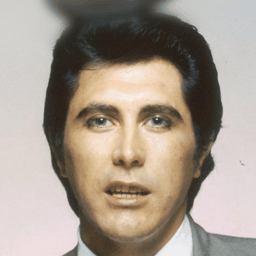}
  \end{tabular}}
  \caption{Qualitative comparison on CelebA -- Gaussian deblurring (noisy).}
  \label{fig:celeba_gauss_noisy_lpips}
\end{figure}
\FloatBarrier
\begin{figure}[h!]
  \setlength{\tabcolsep}{2pt}
  \centering
  \scriptsize
\newcommand{\imgw}{0.12\linewidth}
\newcommand{\imgrow}[7]{
  \includegraphics[width=\imgw]{#1} &
  \includegraphics[width=\imgw]{#2} &
  \includegraphics[width=\imgw]{#3} &
  \includegraphics[width=\imgw]{#4} &
  \includegraphics[width=\imgw]{#5} &
  \includegraphics[width=\imgw]{#6} &
  \includegraphics[width=\imgw]{#7} \\
}
  \resizebox{0.85\linewidth}{!}{
  \begin{tabular}{@{}ccccccc@{}}
    Measured & DPS & DiffPIR & DDRM & DiffPIR+LAMP & DDRM+LAMP & Original \\
    \imgrow{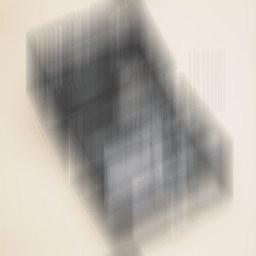}{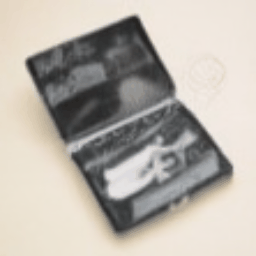}{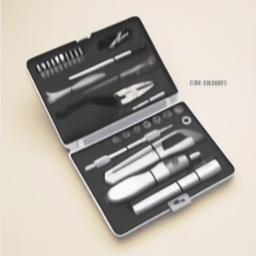}{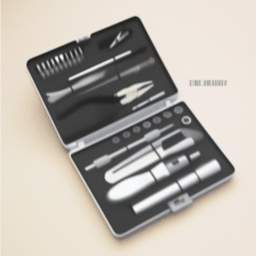}{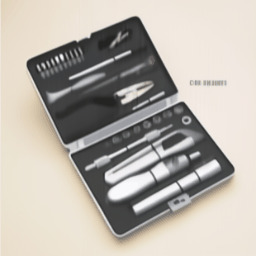}{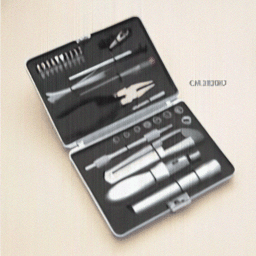}{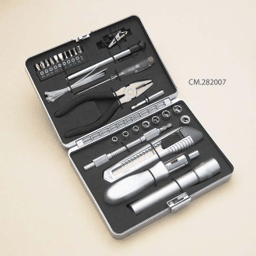}
    \imgrow{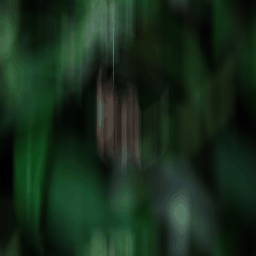}{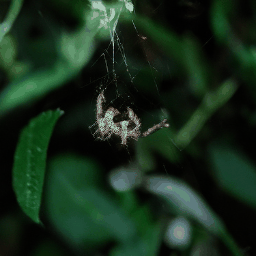}{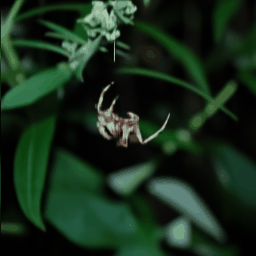}{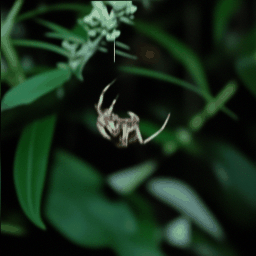}{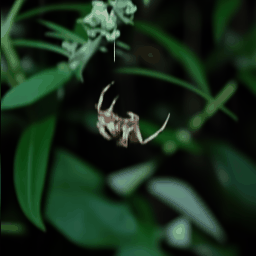}{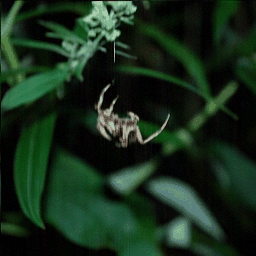}{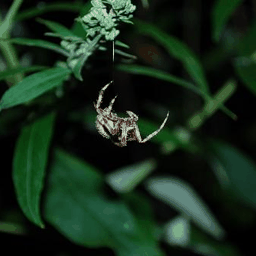}
    \imgrow{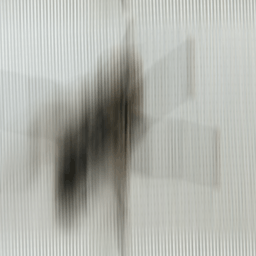}{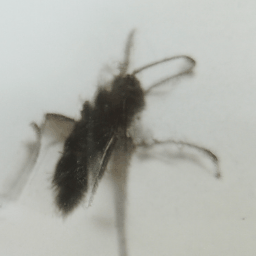}{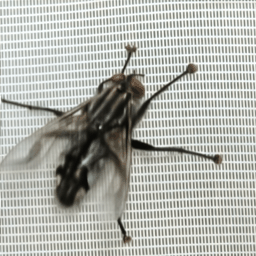}{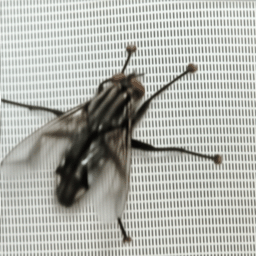}{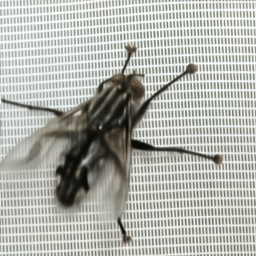}{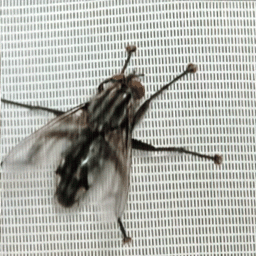}{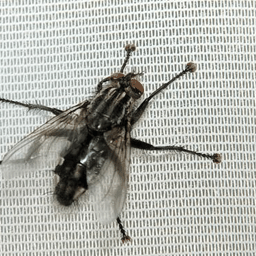}
    \imgrow{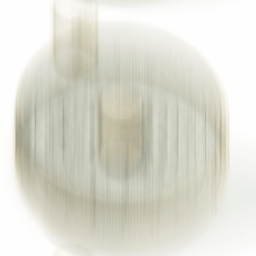}{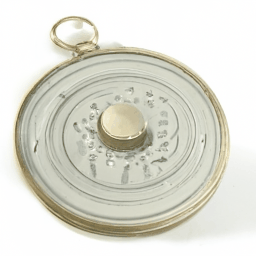}{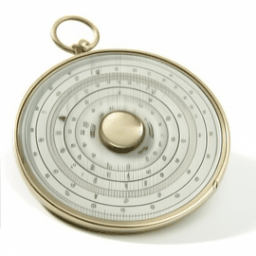}{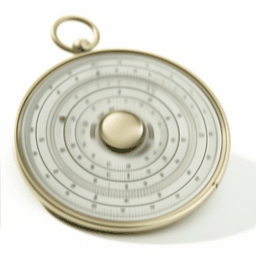}{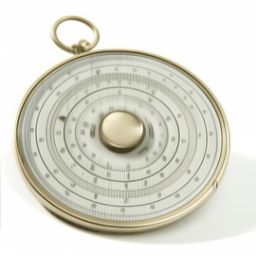}{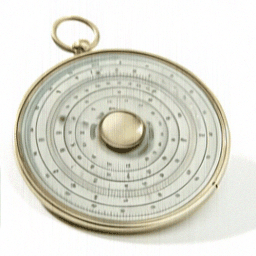}{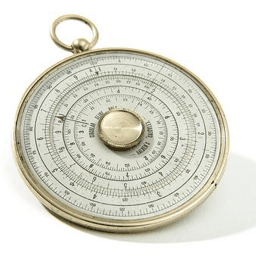}
    \imgrow{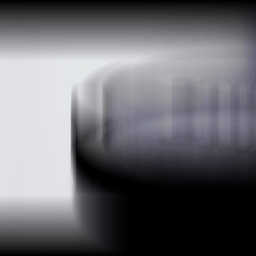}{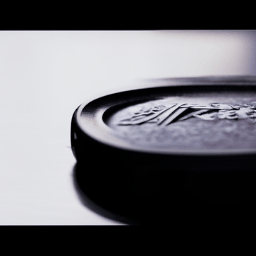}{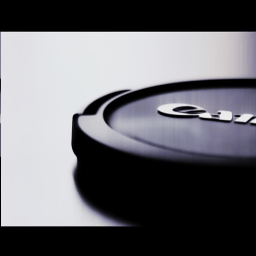}{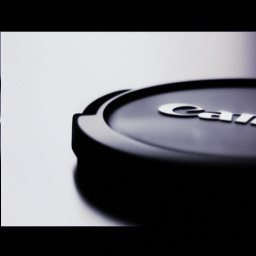}{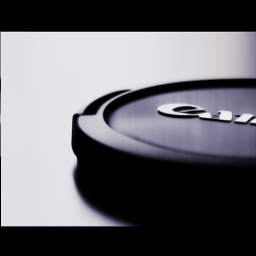}{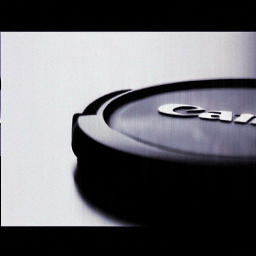}{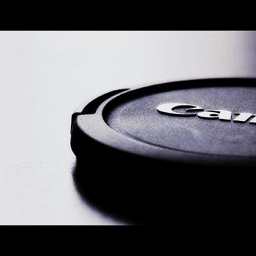}
  \end{tabular}}
  \caption{Qualitative comparison on ImageNet -- Motion deblurring (noiseless).}
  \label{fig:imagenet_motion_noiseless_lpips}
\end{figure}
\FloatBarrier
\begin{figure}[h!]
  \setlength{\tabcolsep}{2pt}
  \centering
  \scriptsize
\newcommand{\imgw}{0.12\linewidth}
\newcommand{\imgrow}[7]{
  \includegraphics[width=\imgw]{#1} &
  \includegraphics[width=\imgw]{#2} &
  \includegraphics[width=\imgw]{#3} &
  \includegraphics[width=\imgw]{#4} &
  \includegraphics[width=\imgw]{#5} &
  \includegraphics[width=\imgw]{#6} &
  \includegraphics[width=\imgw]{#7} \\
}
  \resizebox{0.85\linewidth}{!}{
  \begin{tabular}{@{}ccccccc@{}}
    Measured & DPS & DiffPIR & DDRM & DiffPIR+LAMP & DDRM+LAMP & Original \\
    \imgrow{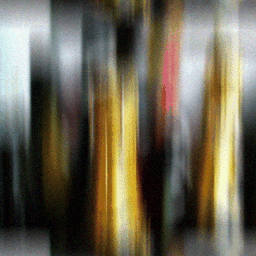}{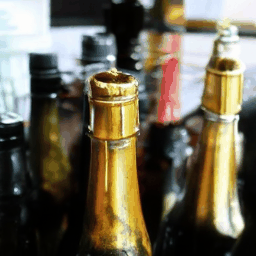}{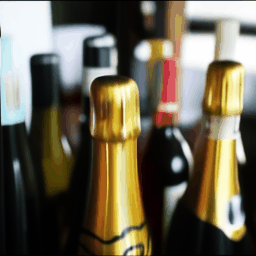}{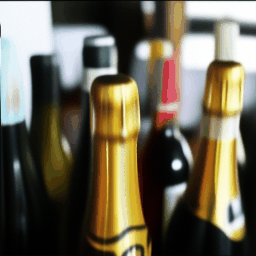}{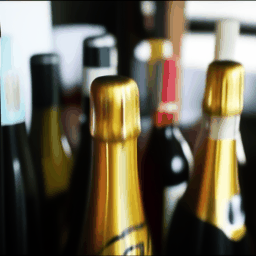}{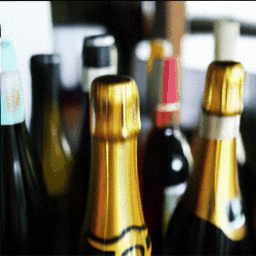}{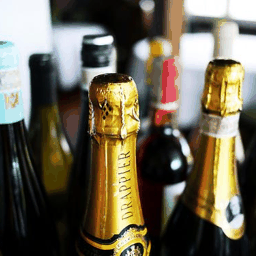}
    \imgrow{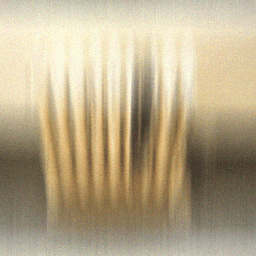}{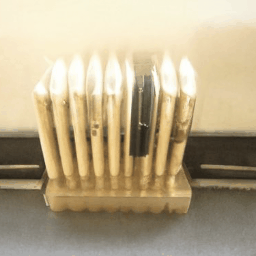}{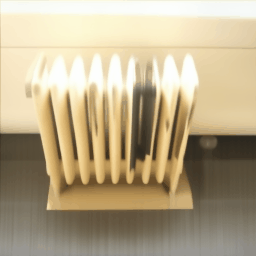}{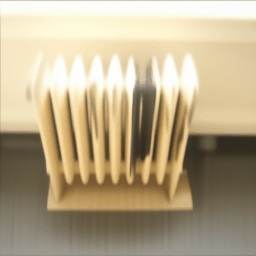}{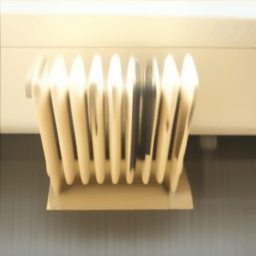}{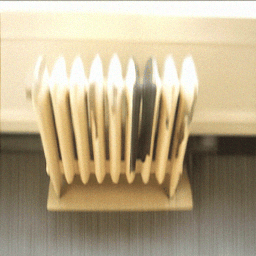}{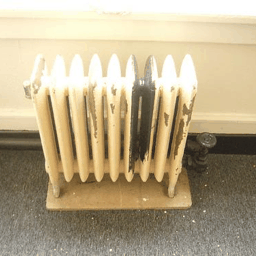}
    \imgrow{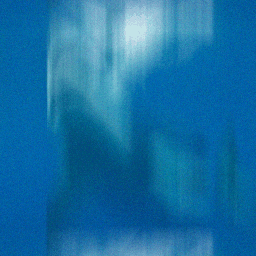}{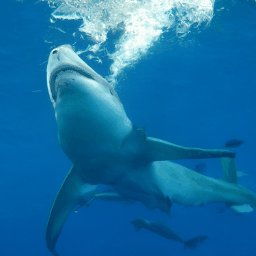}{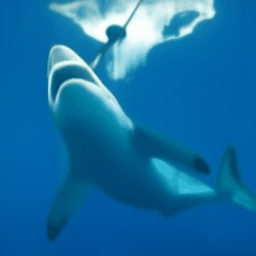}{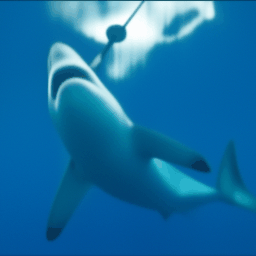}{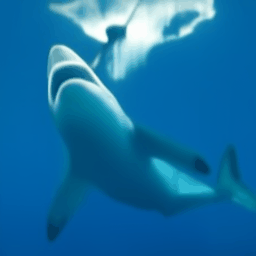}{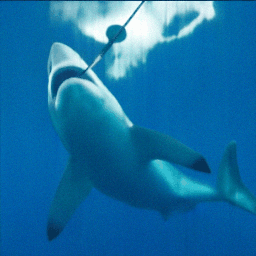}{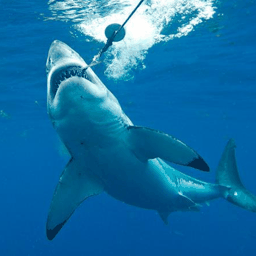}
    \imgrow{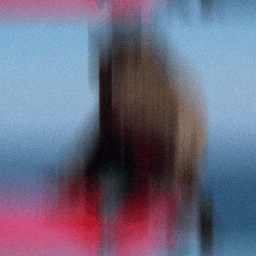}{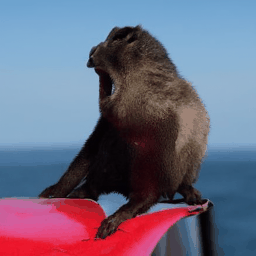}{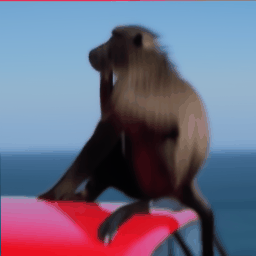}{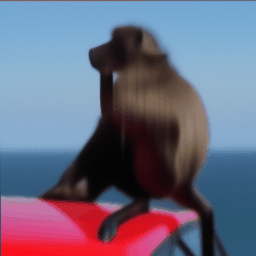}{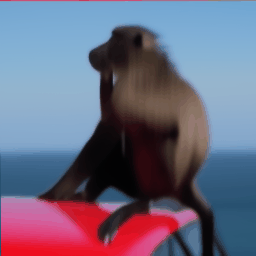}{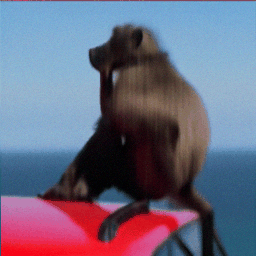}{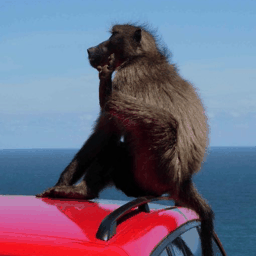}
    \imgrow{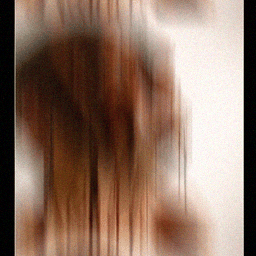}{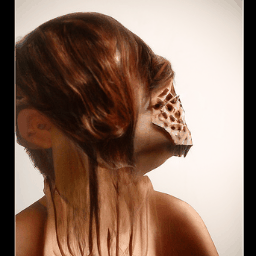}{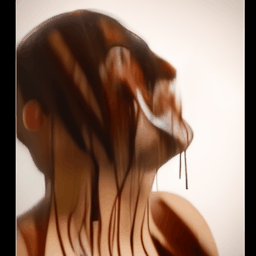}{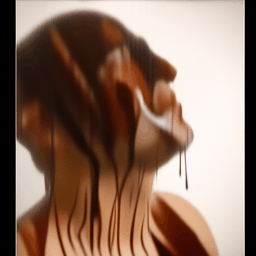}{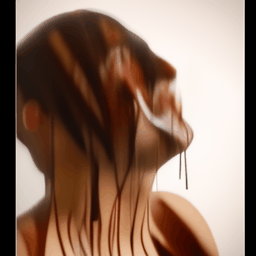}{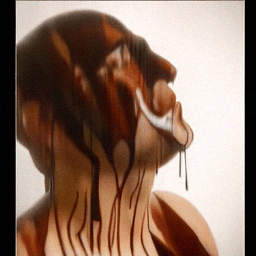}{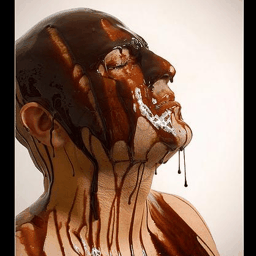}
  \end{tabular}}
  \caption{Qualitative comparison on ImageNet -- Motion deblurring (noisy).}
  \label{fig:imagenet_motion_noisy_lpips}
\end{figure}
\FloatBarrier
\begin{figure}[h!]
  \setlength{\tabcolsep}{2pt}
  \centering
  \scriptsize
\newcommand{\imgw}{0.12\linewidth}
\newcommand{\imgrow}[7]{
  \includegraphics[width=\imgw]{#1} &
  \includegraphics[width=\imgw]{#2} &
  \includegraphics[width=\imgw]{#3} &
  \includegraphics[width=\imgw]{#4} &
  \includegraphics[width=\imgw]{#5} &
  \includegraphics[width=\imgw]{#6} &
  \includegraphics[width=\imgw]{#7} \\
}
  \resizebox{0.85\linewidth}{!}{
  \begin{tabular}{@{}ccccccc@{}}
    Measured & DPS & DiffPIR & DDRM & DiffPIR+LAMP & DDRM+LAMP & Original \\
    \imgrow{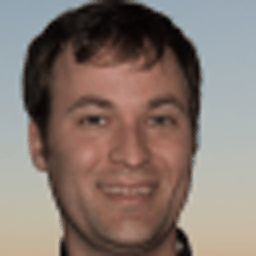}{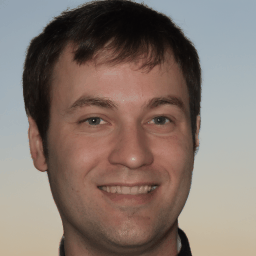}{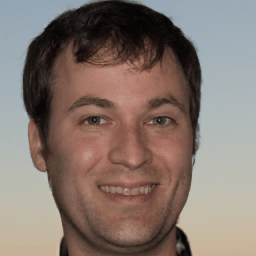}{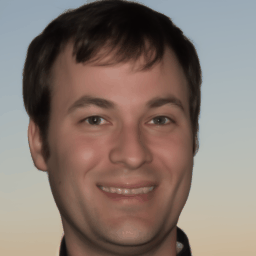}{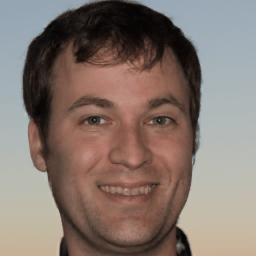}{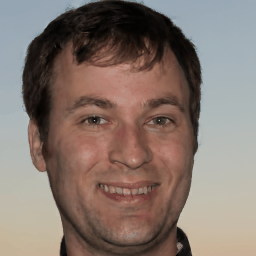}{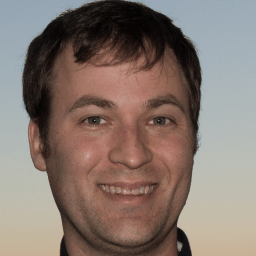}
    \imgrow{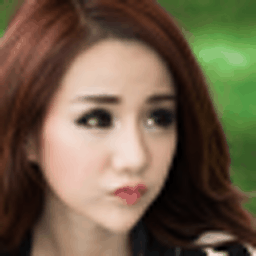}{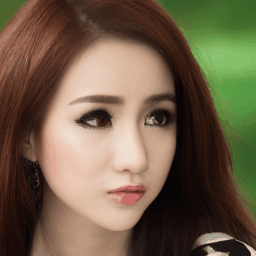}{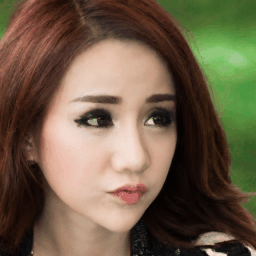}{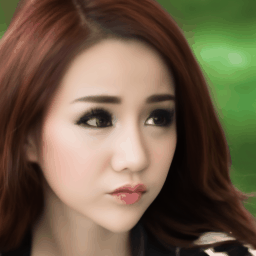}{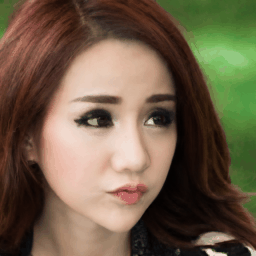}{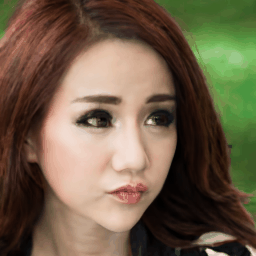}{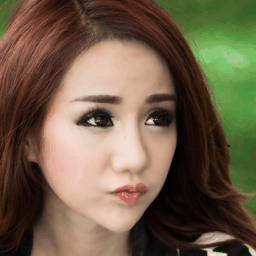}
    \imgrow{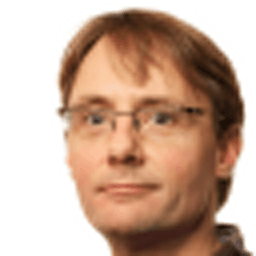}{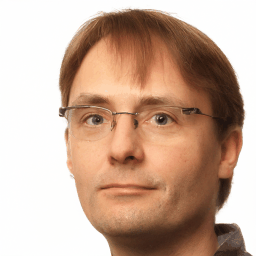}{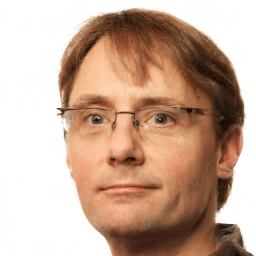}{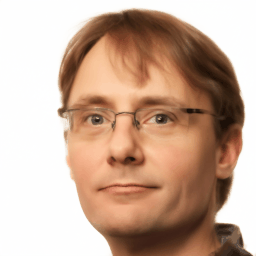}{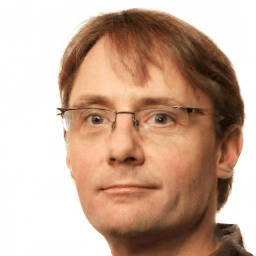}{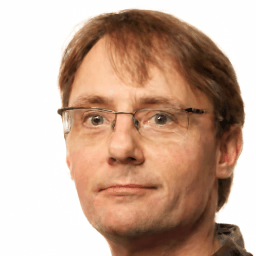}{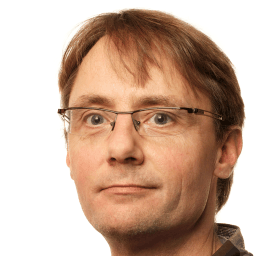}
    \imgrow{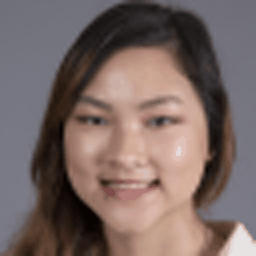}{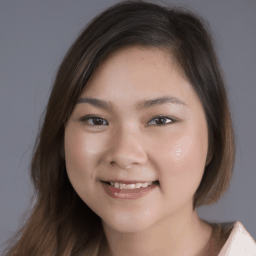}{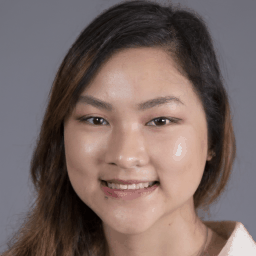}{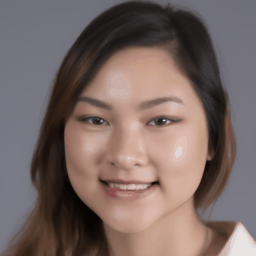}{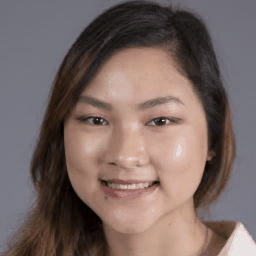}{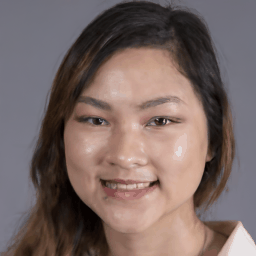}{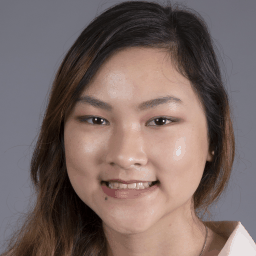}
    \imgrow{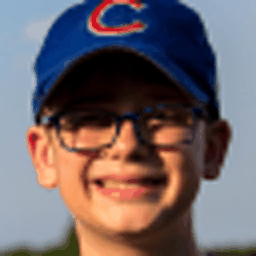}{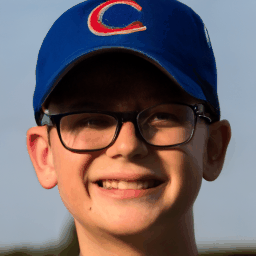}{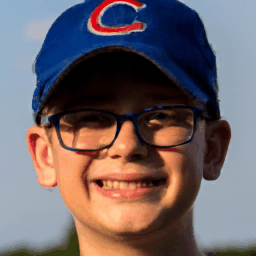}{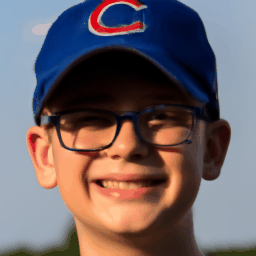}{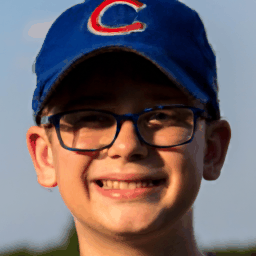}{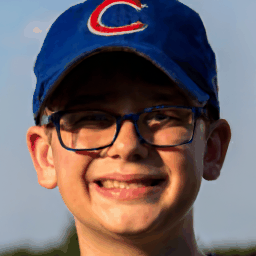}{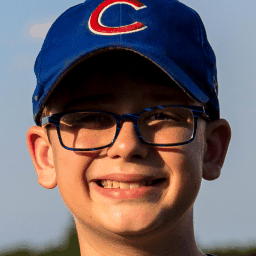}
  \end{tabular}}
  \caption{Qualitative comparison on FFHQ -- $\times 4$ super-resolution (noiseless).}
  \label{fig:ffhq_sr_noiseless_lpips}
\end{figure}
\FloatBarrier
\begin{figure}[h!]
  \setlength{\tabcolsep}{2pt}
  \centering
  \scriptsize
\newcommand{\imgw}{0.12\linewidth}
\newcommand{\imgrow}[7]{
  \includegraphics[width=\imgw]{#1} &
  \includegraphics[width=\imgw]{#2} &
  \includegraphics[width=\imgw]{#3} &
  \includegraphics[width=\imgw]{#4} &
  \includegraphics[width=\imgw]{#5} &
  \includegraphics[width=\imgw]{#6} &
  \includegraphics[width=\imgw]{#7} \\
}
  \resizebox{0.85\linewidth}{!}{
  \begin{tabular}{@{}ccccccc@{}}
    Measured & DPS & DiffPIR & DDRM & DiffPIR+LAMP & DDRM+LAMP & Original \\
    \imgrow{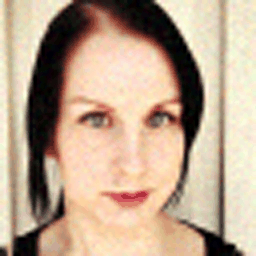}{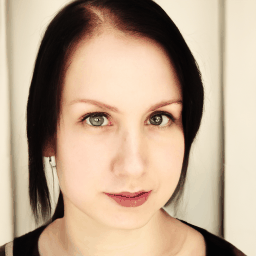}{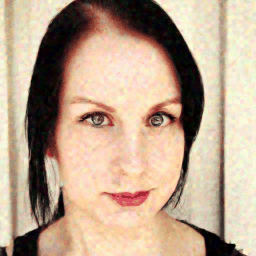}{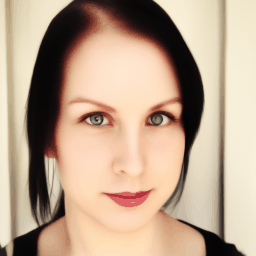}{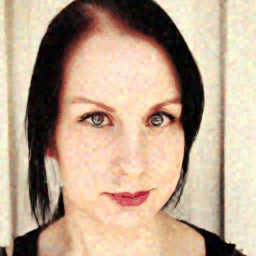}{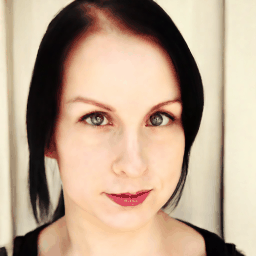}{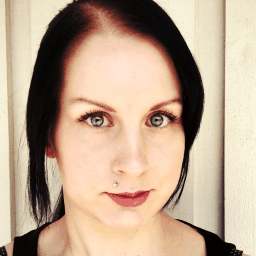}
    \imgrow{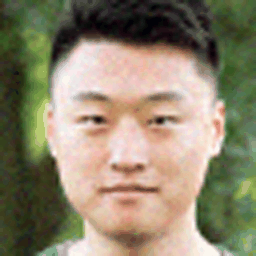}{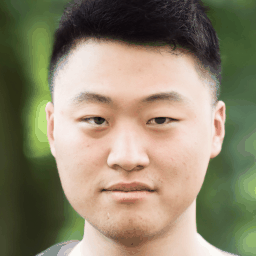}{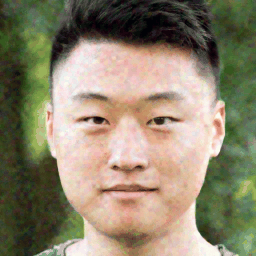}{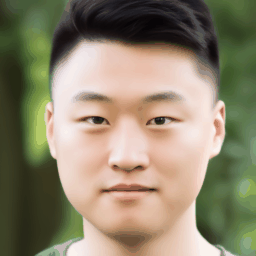}{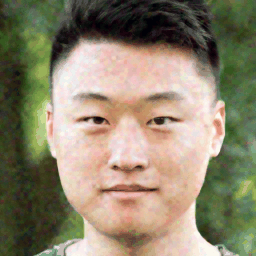}{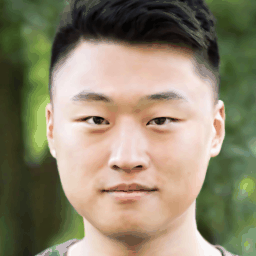}{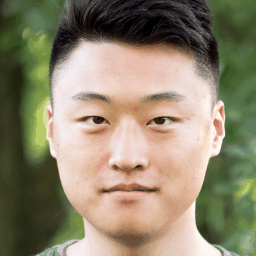}
    \imgrow{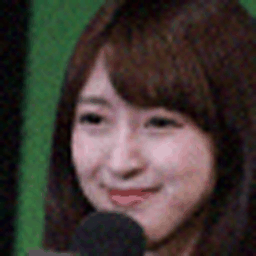}{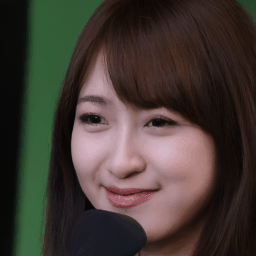}{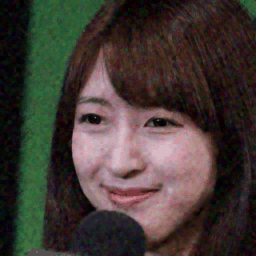}{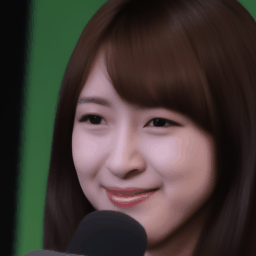}{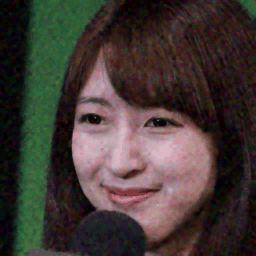}{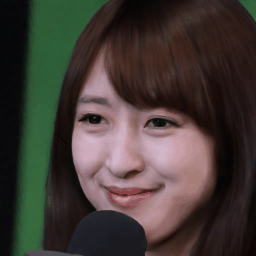}{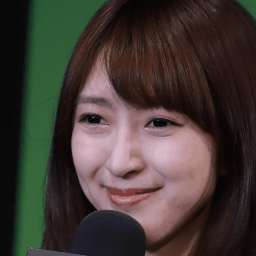}
    \imgrow{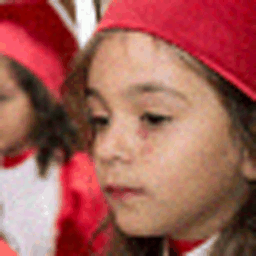}{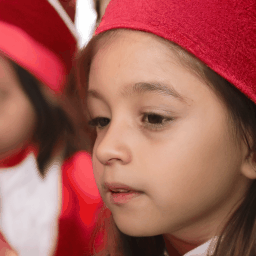}{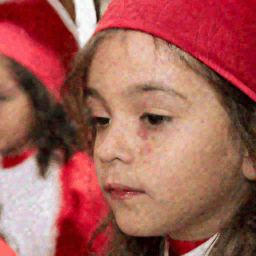}{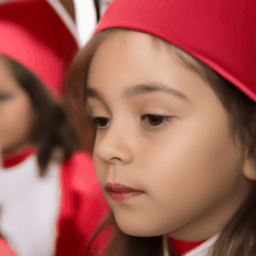}{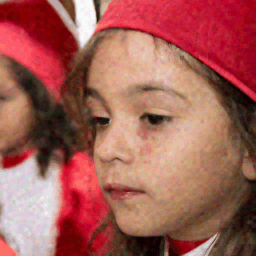}{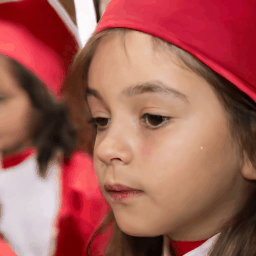}{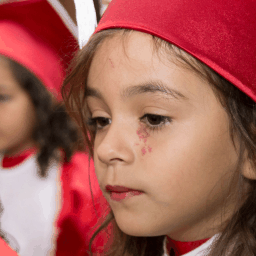}
    \imgrow{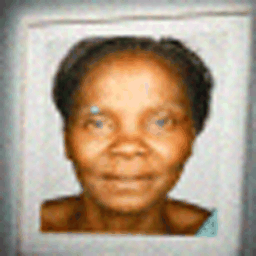}{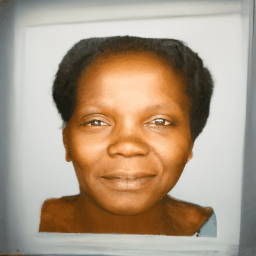}{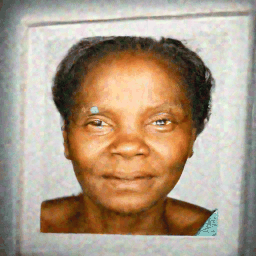}{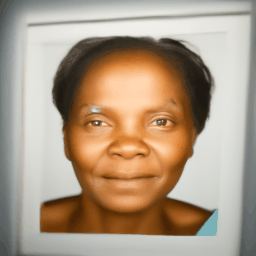}{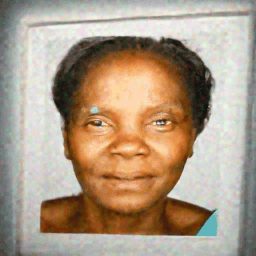}{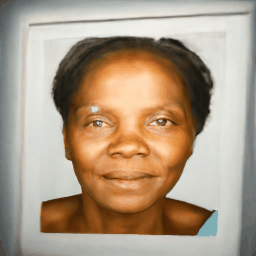}{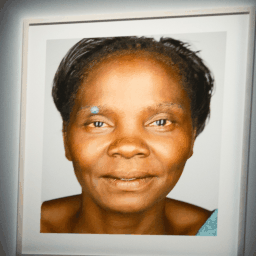}
  \end{tabular}}
  \caption{Qualitative comparison on FFHQ -- $\times 4$ super-resolution (noisy).}
  \label{fig:ffhq_sr_noisy_lpips}
\end{figure}
\FloatBarrier

\end{document}